%% file: sbr_main.tex
\documentclass[runningheads]{llncs}
\usepackage{graphicx}
\usepackage{comment}
\usepackage{amsmath,amssymb,bm} 
\usepackage{color}
\usepackage{xcolor}
\usepackage{lipsum}
\usepackage{tabularx}
\usepackage{adjustbox}
\usepackage{multirow}
\usepackage{siunitx}
\usepackage{dsfont}
\usepackage{xcolor}
\usepackage{array, booktabs, bigdelim, makecell}
\usepackage{subcaption}
\usepackage{epstopdf} 

\newcolumntype{C}[1]{>{\centering\arraybackslash}p{#1}}
\newcolumntype{R}[1]{>{\raggedleft\arraybackslash}p{#1}}
\newcolumntype{L}[1]{>{\raggedright\arraybackslash}p{#1}}

\begin{document}
\pagestyle{headings}
\mainmatter

\title{Sample-based Regularization: A Transfer Learning Strategy Toward Better Generalization} 

\titlerunning{Sample-based Regularization}
%
\author{
Yunho Jeon\inst{1} \and 
Yongseok Choi\inst{2} \and 
Jaesun Park\inst{2}  \and
Subin Yi\inst{2} \and 
Dong-Yeon Cho\inst{2} \and
Jiwon Kim\inst{2} 
}
%
\authorrunning{Y. Jeon et al.}
%
\institute{
mofl inc. \footnote{Work done while being at SK Telecom}\\
\email{jyh2986@gmail.com} \and
Vision AI Labs, SK Telecom\\
\email{\{yongseokchoi, jayden\_park, yisubin, dongyeon.cho, ai.kim\}@sktelecom.com}}
\maketitle

\begin{abstract}
Training a deep neural network with a small amount of data is a challenging problem as it is vulnerable to overfitting. However, one of the practical difficulties that we often face is to collect many samples. Transfer learning is a cost-effective solution to this problem. By using the source model trained with a large-scale dataset, the target model can alleviate the overfitting originated from the lack of training data. Resorting to the ability of generalization of the source model, several methods proposed to use the source knowledge during the whole training procedure. However, this is likely to restrict the potential of the target model and some transferred knowledge from the source can interfere with the training procedure. For improving the generalization performance of the target model with a few training samples, we proposed a regularization method called sample-based regularization (SBR), which does not rely on the source's knowledge during training. With SBR, we suggested a new training framework for transfer learning. Experimental results showed that our framework outperformed existing methods in various configurations.

\end{abstract}

\input{01_intro.tex}
\input{02_related_works}

\input{03_approach}

\input{04_experiment}

\input{05_discussions}

\input{06_conclusions}

\input{10_supplement}

\clearpage
%
%
\bibliographystyle{splncs04}
\bibliography{sbr_main}

\end{document}

%% file: 01_intro.tex
\section{Introduction}

In many vision applications, deep learning has shown promising performance. Most of these successes rely on deep neural network architectures and large numbers of training data. However, in many practical cases in the real world, it is hard to collect enough data to prevent overfitting and the performance is not satisfactory when deep neural networks are trained from scratch with a small number of samples. One simple solution to this difficulty is initializing parameters of the network with a network pre-trained on an extremely large dataset and fine-tuning them. Through this approach, we can get a significant performance gain. 

After a network was trained with a large dataset (e.g., ImageNet \cite{DBLP:journals/ijcv/RussakovskyDSKS15/imagenet}), the trained network can be transferred to many target applications that have different domains and tasks. This pre-trained model is accessible and we call this model as source model. In transfer learning, the objective is to learn a target task using this source model and one of the simplest ways to utilize the source model is to initialize parameters of target model by copying the trained parameters from the source model as mentioned above. However, it is noticeable that only parameters of feature extractor are transferred and the parameters of target dependant layers are randomly initialized since there are some differences between the source and target tasks in general.

Inherently, transfer learning is proposed to overcome the lack of training data. He et al. \cite{he2019rethinking} showed that if the data is enough, the advantage of using the pre-trained model diminishes in terms of classification accuracy. Therefore, transfer learning should focus more on data-hunger scenarios. However, these cases are not fully explored and still have limited performance. In the situations where a network should be trained with a small set of images, it is important to increase generalization performance. It is very vulnerable to overfitting with a few training images, and existing methods \cite{l2sp,delta,DBLP:conf/iclr/Wang20/pay-attention-features} also focused on how to regularize the network for increasing generalization performance. These approaches utilized the source model as a reference model and keep the target model as close as the source model. They assumed that the source model had valuable knowledge for the target task and tried to preserve the parameters or behaviors of the source model as much as possible.


However, regularizing with the source model during training can restrict the potential of the target model and some knowledge of the source might be useless for the target task\cite{bss}. Because the objective of transfer learning is to increase the performance of the target task, it is not desirable property to keep the source knowledge as possible for transfer learning. The difference from the source model is acceptable if the performance of the target task can be increased. Regularization using the source model as a reference has been broadly used base on the belief that the source model can generalize better than the target model as it had been trained with a large dataset. In general, however, the target domain and task are different from source configuration. Thus, the retaining source model would be useful only if the domain gap is small. 

In transfer learning, a pre-trained model can be regarded as an initial position for optimization. Therefore, the knowledge of the source task is already transferred to the target model simply using pre-trained weight as an initial point. The next question should be how to improve the performance of the target task from this starting point. Without using the source model as a regularization reference, other pivots are needed for preventing overfitting. From this perspective, we propose to use training samples instead of the source model. In a similar way that knowledge distillation \cite{DBLP:journals/corr/HintonVD15/kd,DBLP:journals/corr/RomeroBKCGB14/fitnet} regularize the output of the target model to follow that of the source model, our method uses the output of training samples of the same class; we called this as sample-based regularization (SBR). Each sample is a regularization reference to other samples in the same class. The target network is regularized to maximize the similarity between samples. In this way, the target model is prevented to overfit even using a small set of training data.

Maximizing similarity can be thought of as metric learning \cite{DBLP:journals/ftml/Kulis13/metric-learning} because it guides the network to make features as close as possible if two samples are in the same class. The difference is our regularization does not separate the samples of a different class. The discrimination between classes is forced by supervised cross-entropy loss. It is known that pairwise similarity is more transferable than discriminate features\cite{DBLP:conf/iclr/HsuLK18/l2c}, which means that training based on a pairwise relationship can be more general on test data rather than training with a one-hot target. 

Based on these findings, we propose an effective framework to increase performance in transfer learning with a small amount of dataset. We divide the target model into the classifier and feature extractor. Classifier focus to separate different classes with cross-entropy loss and feature extractor is forced to learn a similarity of samples with the same class. For this goal, we applied SBR on the output feature of feature extractor and decreased the error propagation to feature extractor from cross-entropy loss to reduce the influence of classifier. Experimental results on various datasets with varying the portion of the training samples outperform existing methods and show the effectiveness of our framework.

%% file: 02_related_works.tex
\section{Related Works}
\noindent\textbf{Transfer Learning}
Transfer learning has been studied for a long time to solve real-world machine learning problems in which only a small amount of training data are available \cite{DBLP:journals/tkde/PanY10/survey-tl}.
In various computer vision tasks, it has been shown that transferable feature representations can be obtained from deep neural networks \cite{DBLP:conf/icml/DonahueJVHZTD14/decaf,DBLP:conf/nips/YosinskiCBL14/dnn-transfer} trained on large-scale datasets such as ImageNet \cite{DBLP:journals/ijcv/RussakovskyDSKS15/imagenet}.
The usual way to utilize these representations is to inherit both the learned parameters and the architecture from the source model and fine-tune on a target dataset while replacing the classifier with target-specific layers.

Recent studies have revealed that the fine-tuning-based transfer can be improved by adding proper regularization to minimize the distance between the source and target parameters \cite{l2sp} or activations \cite{delta,DBLP:conf/iclr/Wang20/pay-attention-features}, or to penalize small singular values of the matrix comprised on feature activations \cite{bss}. 
In \cite{DBLP:conf/iclr/Li20/rethinking-ft}, hyperparameters for fine-tuning have been explored extensively including those of the regularization methods while showing their dependencies on the similarity between the source and target domains.
Our method can be seen as the regularization also, but differs from others in the sense that ours focuses on the relation between different target examples.

Some studies showed that the fine-tuning from ImageNet pre-trained models might not help to improve the final performance of a target task when the target dataset is large enough or a large gap between the source and target domains exists \cite{DBLP:conf/cvpr/KornblithSL19/better-imagenet,he2019rethinking,DBLP:conf/nips/RaghuZKB19/transfusion}.
However, they have observed that the transfer still benefits a speed-up of convergence even in that situation.
One of them has suggested that the faster convergence results from the transfer of weight scaling rather than feature representations \cite{DBLP:conf/nips/RaghuZKB19/transfusion}.



\noindent\textbf{Similarity-based Learning}
Similarity information has been exploited to characterize the relation between two samples in many machine learning tasks including computer vision \cite{DBLP:journals/jmlr/ChenGGRC09/sim-based-cls}.
Regarding the multi-class classification, pairwise similarity is considered more general than categorical information so that the similarity predictor learned in one domain can be transferred to other domains even in the case that the categorical one fails to transfer to new domains.
Based on this idea, one transfer learning scheme has been proposed to train a multi-class classifier only with unlabeled target data by learning to predict the pairwise similarity and cluster with this information in the source dataset \cite{DBLP:conf/iclr/HsuLK18/l2c}. The similarity information can be used together with different types of loss terms for further improvement. To penalize the distances between the feature representations and their corresponding class centers, a new loss term was added to the softmax loss for a face recognition task  \cite{DBLP:conf/eccv/WenZL016/center-loss}. 

\noindent\textbf{Few-shot Learning}
Few-shot learning is related with transfer learning as many methods assume prior knowledge can lead to better generalization in cases that only a small number of training examples is available \cite{DBLP:journals/corr/abs-1904-05046/fsl-survey,DBLP:journals/corr/abs-1808-04572/fsl-survery-2}. 
While meta-learning approach has been popular to solve this problem, recent studies have reported that fine-tuning following conventional pre-training performs similarly to or outperforms the meta-learning \cite{DBLP:conf/iclr/ChenLKWH19/closer-look}. 
However, the conventional few-shot learning evaluation measure assumes that the source (meta-train) and target (meta-test) datasets come from the same domain, and focuses on small problems (e.g., 5 or small number of classes), which are not the cases for practical transfer learning. 
Recently, more realistic benchmarks for few-shot learning have been proposed to overcome these limitations \cite{DBLP:conf/iclr/Triantafillou20/meta-dataset,DBLP:journals/corr/abs-1912-07200/cd-fsl}.








%% file: 03_approach.tex
\section{Approach}
\begin{figure}[!t]
\begin{subfigure}[b]{.6\textwidth}
  \centering
  \includegraphics[width=0.9\linewidth]{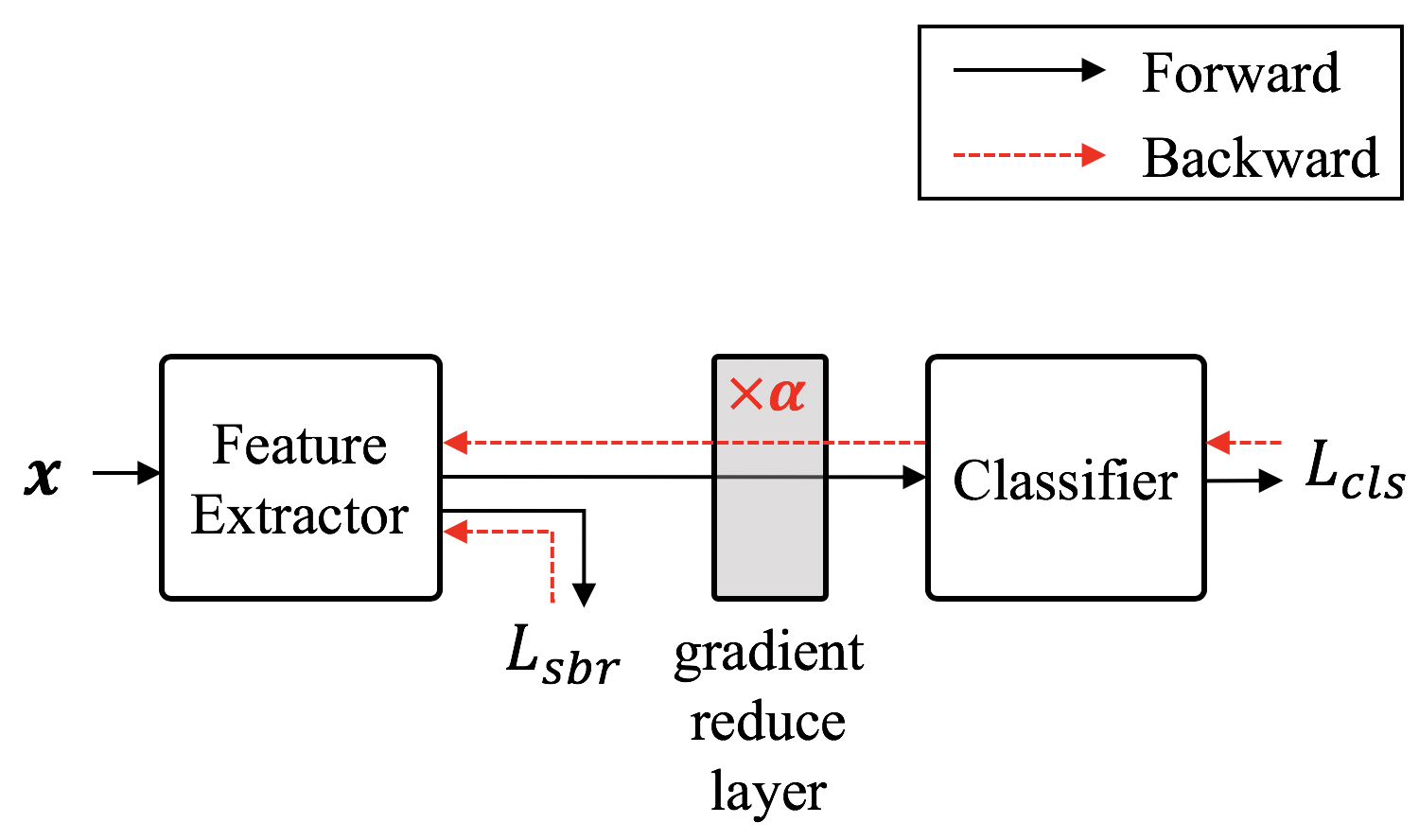}
  \caption{Overall framework}
  \label{fig:overall flow}
\end{subfigure}
\begin{subfigure}[b]{.4\textwidth}
  \centering
  \includegraphics[width=0.8\linewidth]{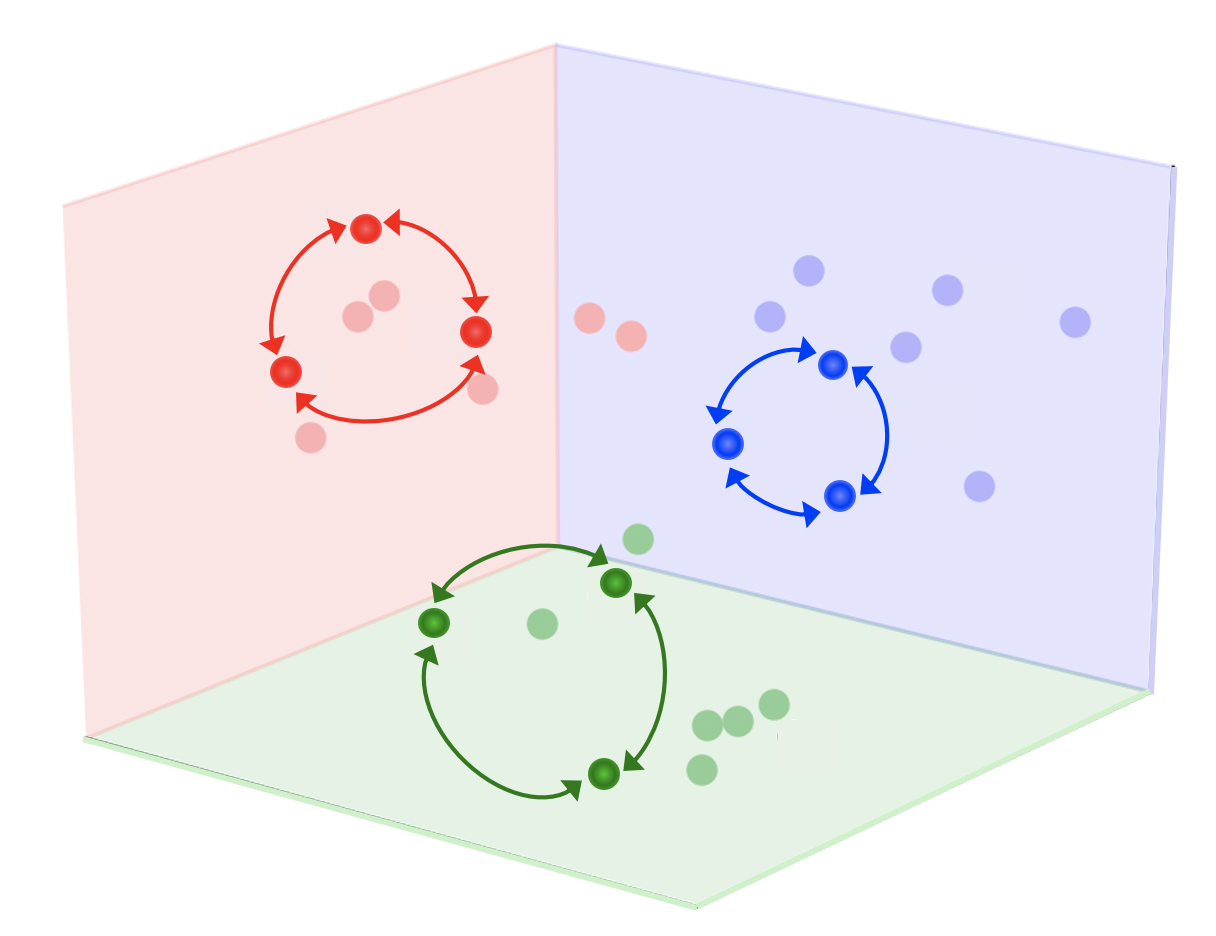}
  \vspace{0.2cm}
  \caption{Sample-based regularization}
  \label{fig:SBR concept}
\end{subfigure}
\caption{(a) Feature extractor is trained by cross-entropy loss ($L_{cls}$) and sample-based regularization ($L_{sbr}$). The graident from classifier is reduced by a ratio of $\alpha$ to weaken the influence of cross-entropy for better generalization. (b) Each color represents different class and SBR encourages the samples within the same class to be similar to each other.}
\label{fig:concept}
\end{figure}

In this section, the proposed transfer learning algorithm with sample-based regularization (SBR) is explained. Fig. \ref{fig:concept}(a) shows the overall flow of our training procedure where the target model is divided into the classifier and feature extractor. Cross-entropy loss with one-hot labels is used for training the classifier as in conventional supervised training. Feature extractor also trained with back-propagated error from the classifier but its gradient is reduced by a certain ratio to weaken the influence of cross-entropy loss. Instead, we apply SBR to the feature extractor for better generalization performance.

\subsection{Background}
Given labeled training dataset $\mathbf{X} = \{(x_i, y_i)\}^N_{i=1}$, supervised learning with a deep neural network model can be formulated as follows: 
\begin{align}
  \underset{\mathbf{\mathbf}}{\min} \sum_{i=1}^{N} L(g(f(x_i, \mathbf{w}_{f}), \mathbf{w}_g), y_i) + \lambda \Omega(\mathbf{w}, \cdot),
\end{align}
where \textit{g} and \textit{f} refer to the classifier and the feature extractor of a target model, respectively. \textbf{w} is weight parameters of the model, which is the union of classifier parameter $\textbf{w}_g$ and feature extractor parameter $\textbf{w}_f$. $\Omega(\mathbf{w}, \cdot)$ is a regularization term to prevent overfitting (e.g., $\Omega = ||\textbf{w}||_2^2$, if L2 regularization is applied). In transfer learning, feature extractor \textit{f} has same architecture with the source model and $\mathbf{w}_{f}$ is initialized with pre-trained weight $\mathbf{w}^*_{f}$. As the classifier \textit{g} should complete a target-specific task, it is usually different from the source model and $\mathbf{w}_{g}$ is randomly initialized.

To increase the generalization performance, existing methods proposed different types of regularization. L2-SP \cite{l2sp} proposed to use $\Omega(\textbf{w}) = \alpha ||\textbf{w}_f - \textbf{w}^*_f||_2^2 + \beta ||\textbf{w}_g||_2^2$ to keep the parameters of the target model as close to that of the source model. In DELTA \cite{delta}, instead of regularizing the parameters of the target model, they regularized the behavior of the target model by keeping the output of the target model similar with the source model as follows:

\begin{align}
\Omega(\textbf{w}, \cdot) = \sum_{i=1}^{N}\sum_j A_j(\cdot) ||FM_j(f, x_i, \textbf{w}_f) - FM_j(f, x_i, \textbf{w}^*_f)||_2^2 + \beta ||\textbf{w}_g||_2^2,
\end{align}
where $FM_j$ is the output feature of $j_{th}$ filter in feature extractor and $A_j(\cdot)$ is an attention function conditioned on the input and the source model. The regularization of DELTA method is similar to knowledge distillation (KD) \cite{DBLP:journals/corr/HintonVD15/kd} between the source feature and the target feature. The distinction is that KD distills knowledge within the same task and domain but DELTA considers different ones. Beyond these methods, batch spectral shrinkage (BSS) \cite{bss} pointed that transferring the source model can cause negative transfer which disturbs the training of the target model by transferring unwanted knowledge. They alleviated this problem by penalizing smaller singular values of the feature matrix from feature extractor instead of regularizing parameters directly.


\subsection{Sample-based Regularization}
Many existing methods apply regularization based on the source model for fine-tuning. These regularization methods rely on the ability of the generalization performance of the source. Differently from KD, however, the target task and domain are not same as the source in transfer learning. As pointed in \cite{bss}, not all knowledge is transferable from the source and it might give a negative effect for training the target model. However, as the objective of the transfer learning is to increase the performance of the target task, a regularization based on the source model is not essential.

Instead of using the source model, we propose to use outputs of other samples within the same class as a reference feature. This procedure can be regarded as sample-based distillation. If KD distills the output feature of feature extractor \textit{f}, the regularization of KD is formulated as follows:
\begin{align}
\label{eq:KD}
\Omega(\textbf{w}, \cdot) = \sum_{i=1}^{N} \textbf{D}(f(x_i, \textbf{w}^*_f), f(x_i, \textbf{w}_f)),
\end{align}
where \textbf{D} is a measure of dissimilarity between two features. For each sample $x_i$, KD makes two features from different models close; this makes sense as the task and domain is the same. As this assumption is not satisfied in the transfer learning, our sample-based regularization is defined as follows:
\begin{align}
\label{eq:Sample Based KD}
\Omega(\textbf{w}, \cdot) = \sum_{c=1}^{C} \sum_{(x_i,x_j) \in \textbf{X}_c} \textbf{D}(f(x_i, \textbf{w}_f), f(x_j, \textbf{w}_f)),
\end{align}
where $C$ is the number of classes and $\textbf{X}_c$ is the set of samples within the same class (i.e. $\textbf{X}_c = {\{(x_i,x_j)|y_i=c, y_j=c\}}$). After initializing $w_f$ with $\textbf{w}^*_f$ in the target model, we use different samples within the same class considering only the target model to distill knowledge to each other. As the target model has been initialized with the pre-trained model, we hypothesize that sample features from the target model have already some useful information from the source model.

Calculating all pairs of samples within the same class in the entire training dataset is time-consuming and impractical. Instead, we use a stochastic update to calculate the regularization term in Eq. (\ref{eq:Sample Based KD}) and the loss for this sample-based regularization is defined as follows: 

\begin{align}
\label{eq:pairwise_distance}
  L_{sbr} = \sum_{c=1}^C {1\over N^{pair}_c }\sum_{(x_i,x_j) \in B_c} \textbf{D}(f(x_i, \textbf{w}_f) , f(x_j,  \textbf{w}_f)),
\end{align}
where $N^{pair}_c$ is a normalization factor and equals to the number of pairs in the same class \textit{c} (i.e., $N^{pair}_c = {N_c}(N_c - 1)$, $N_c$ is the number of class-$c$ samples within a batch; all permutation of pairs are counted. If \textbf{D} is symmetric, the same values are summed twice). $B_c$ is the class-\textit{c} samples within a mini-batch. $L_{sbr}$ is calculated for each mini-batch and the error is backpropagated to the feature extractor $f$ in the target model. Fig. \ref{fig:concept}(b) shows the concept of SBR. Among many samples in the training data, SBR is applied to samples in a batch.

\subsection{Reducing Influence of Cross-entropy Loss}
Cross-entropy with softmax loss commonly has been used in supervised training and it works well for finding decision boundaries between classes. However, as it matches class probabilities to one-hot labels, it does not care about the distribution of features and sometimes causes over-confidence problem \cite{guo17a}. If the training data is large enough, the true distribution of features is not that different from training distribution and the generalization performance of the trained model would be fine. However, with a small number of training data, the distribution of features is sparse and many local optima for deciding decision boundaries exist. In this situation, learning only with cross-entropy loss can cause a overfitting problem. 

If the decision boundary that the classifier decided is fixed too early, the back-propagated gradient to the feature extractor might not be useful. 
This problem can be alleviated by using adjusting learning rate or momentum if domain difference between the source and the target is not large \cite{DBLP:conf/iclr/Li20/rethinking-ft}. However, fundamentally this problem still exists as cross-entropy loss separates feature space only, not improving the generality of features. 

Recent papers showed that pairwise relationship can improve generalization performance of extracted features \cite{DBLP:conf/iclr/HsuLK18/l2c}. As SBR is also considering relationship of samples, it helps to increase the performance with a small amount of data and we can consider using SBR as a major loss for feature extractor. However, if we use SBR loss only, the extracted feature can collapse to one feature because there is no explicit signal for class discrimination.

In view of these matters, we propose the following training framework for transfer learning:

\begin{align}
  L_{g} &= L_{cls} + \Omega(\mathbf{w}, \cdot), \\
  L_{f} &= \alpha L_{cls} + \beta L_{sbr} + \lambda \Omega(\mathbf{w}, \cdot),
\end{align}
where $L_{g}$ and $L_{f}$ are losses for the classifier and feature extractor respectively, and $L_{cls}$ is a general cross-entropy loss for a classification task. $\Omega$ is a regularization term for weights and L2 regularization is used generally. $\alpha$, $\beta$ and $\lambda$ are hyperparameters for controlling the strength of each part. The classifier is trained with standard training procedure and the feature extractor is trained with two losses. To weaken the influence of the cross-entropy loss, we reduced gradient from $L_{cls}$ to feature extractor by the ratio of $\alpha$ ($0<\alpha \le 1$). Fig. \ref{fig:concept}(a) shows the proposed framework. The gradient-reduce layer does not change the input in the forward pass but reduces the gradient from classifier with the ratio of $\alpha$ in the backward pass.




%% file: 04_experiment.tex
\section{Experiments}
We tested the proposed framework on various datasets for transfer learning including CUB-200-2011 \cite{cub200}, FGVC Aircraft \cite{aircraft}, Stanford Cars \cite{cars}, Stanford Dogs \cite{dogs}, and Oxford Flowers \cite{flowers}. These datasets have different numbers of training examples and classes. Table \ref{table:dataset} summarizes the datasets used in our experiments.

For the source model, we used ImageNet-1k \cite{DBLP:journals/ijcv/RussakovskyDSKS15/imagenet} and Places365 \cite{places} datasets to check the generality of our method regarding the domain difference between the source and the target. Furthermore, we also evaluated accuracy by using only 50\%, 30\% and 15\% of the training examples to check the robustness of our method in cases that a limited number of training samples is available.

\begin{table}[!t]
\begin{center}
\caption{Summary of datasets. The number of samples for each split and classes. Train/class is the average number of training samples per class. For the Aircraft and Flowers dataset, we combined the original train and validation splits following \cite{DBLP:conf/iclr/Li20/rethinking-ft}.}
\label{table:dataset}
\begin{tabular}{C{1.8cm}L{3cm}R{1.3cm}R{1.3cm}R{1.3cm}R{1.8cm}}
\hline\noalign{\smallskip}
& Dataset                                                          & Train        & Test         & Classes & Train/class \\ 
\noalign{\smallskip}
\hline
\noalign{\smallskip}
\multirow{2}{*}{Source}&ImageNet-1k \cite{DBLP:journals/ijcv/RussakovskyDSKS15/imagenet} & 1,281,167    & 100,000      & 1,000   & 1281.17  \\
&Places365 \cite{places}                                          & 1,803,460    & 36,500       & 365     & 4940.99  \\
\hline
\noalign{\smallskip}
\multirow{5}{*}{Target} & CUB-200-2011 \cite{cub200}                                       & 5,994        &  5,794       & 200     &   29.97  \\
&FGVC Aircraft \cite{aircraft}                                    & 6,667        &  3,333       & 100     &   66.67  \\
&Stanford Cars \cite{cars}                                        & 8,144        &  8,041       & 196     &   41.55  \\
&Stanford Dogs \cite{dogs}                                        & 12,000       &  8,580       & 120     &  100.00  \\
&Oxford Flowers \cite{flowers}                                    & 2,040        &  6,149       & 102       & 20.00  \\
\hline
\end{tabular}
\end{center}
\end{table}

ResNet-50 \cite{DBLP:conf/cvpr/HeZRS16_resNet} is used for experiment. We adjusted the base learning rate according to the source and target datasets as the best learning rate depends on those \cite{DBLP:conf/iclr/Li20/rethinking-ft}. We trained models for 200 epoch by using SGD with cosine annealing of the learning rate \cite{DBLP:conf/iclr/LoshchilovH17}. We set the learning rate for the feature extractor 10 times smaller than the classifier following the convention of transfer learning. We denote the learning rate of the classifier as default in this paper. For our method, we set $\alpha$ to 0.1 and we do not reduce the learning rate for the feature extractor to match the magnitude of gradient from the classifier. The detailed meaning of this configuration is analyzed in section \ref{sec:diff_gr_fe}. 

For the dissimilarity measure \textbf{D}, we used squared Euclidean distance $D(\textbf{a}, \textbf{b}) = {1\over 2}||\textbf{a} - \textbf{b}||^2$. Using this measure not only performs well but has a good property for calculation (see section \ref{sec:SBR with L2}) and we also compared with using different measures (section \ref{sec:exp_diff_similarity}). $\beta$ depends on the source and the target data. We found an appropriate scale of $\beta$ for each dataset and kept the same $\beta$ while changing the portion of datasets. However, the best $\beta$ can be different depending on the size of training examples; this was also analyzed in section \ref{sec:beta vs samples}.

\begin{table}[!t]
\begin{center}
\caption{Improvement (\%) of the test accuracy from the baseline.  From ImageNet pre-trained model is used as a source. Our method outperforms other methods on various configurations and datasets. BSS Best represents the best accuracy of BSS among three combinations (L2+BSS, L2-SP+BSS, DELTA+BSS) and these values are referenced from  \cite{bss}}
\label{table:main_exp_imagenet}
\begin{tabular}{C{1.8cm}C{1.8cm}R{1.8cm}R{1.8cm}R{1.8cm}R{1.8cm}}
\hline\noalign{\smallskip}
\multirow{2}{*}{Dataset} & \multirow{2}{*}{Method} & \multicolumn{4}{c}{Sampling Rate}\\
            \cmidrule(l){3-6}    &  & \multicolumn{1}{c}{15\%} & \multicolumn{1}{c}{30\%} & \multicolumn{1}{c}{50\%} & \multicolumn{1}{c}{100\%} \\
\noalign{\smallskip}
\hline
\hline
\noalign{\smallskip}
\multirow{4}{*}{CUB-200}   
&  L2-SP\cite{l2sp}      &0.14$\pm$0.51 & 0.35$\pm$0.14 & -0.12$\pm$0.24& -0.40$\pm$0.30 \\
&  DELTA\cite{delta}     &0.62$\pm$0.45 & 0.52$\pm$0.39 & 0.03$\pm$0.57 & 0.12$\pm$0.33 \\
&  BSS Best\cite{bss}    &4.52$\pm$0.07 & 3.7$\pm$0.29 & 2.44$\pm$0.17 & 1.35$\pm$0.12 \\
&  SBR(ours) &\textbf{13.69$\pm$0.67} & \textbf{8.78$\pm$0.24} & \textbf{4.82$\pm$0.19} & \textbf{2.79$\pm$0.15} \\
\hline

\noalign{\medskip}
\multirow{4}{*}{Dogs}
&  L2-SP      &-0.15$\pm$0.34 & 0.25$\pm$0.24 & 0.23$\pm$0.07 & 0.45$\pm$0.18 \\
&  DELTA     &1.49$\pm$2.67 & -0.15$\pm$0.04 & 0.66$\pm$1.03 & 0.05$\pm$0.13 \\
&  BSS Best  &1.15$\pm$0.27 & 0.59$\pm$0.17 & 0.49$\pm$0.05 & 0.29$\pm$0.14 \\
&  SBR(ours) &\textbf{5.99$\pm$0.02} & \textbf{3.34$\pm$0.12} & \textbf{1.97$\pm$0.04} & \textbf{1.32$\pm$0.10} \\
\hline

\noalign{\medskip}
\multirow{4}{*}{Cars}
&  L2-SP      &0.61$\pm$034 & 0.83$\pm$0.40 & 0.76$\pm$0.25 & -0.15$\pm$0.45 \\
&  DELTA     &0.51$\pm$0.99 & -0.23$\pm$0.19 & 0.31$\pm$0.12 & -0.19$\pm$0.27 \\
&  BSS Best  &5.15$\pm$0.16 & 4.04$\pm$0.28 & 2.48$\pm$0.33 & 0.43$\pm$0.27 \\
&  SBR(ours) &\textbf{12.54$\pm$1.11} & \textbf{10.94$\pm$0.66} & \textbf{6.85$\pm$0.26} & \textbf{2.28$\pm$0.10} \\
\hline

\noalign{\medskip}
\multirow{4}{*}{Aircraft}
&  L2-SP      &0.24$\pm$0.60 & 0.07$\pm$1.37 & 0.25$\pm$0.62 & -0.20$\pm$0.30 \\
&  DELTA     &0.55$\pm$0.73 & 0.43$\pm$0.77 & 0.10$\pm$0.32 & -0.35$\pm$0.45 \\
&  BSS Best  &4.22$\pm$0.19 & 4.12$\pm$0.17 & 1.53$\pm$0.29 & 0.35$\pm$0.18 \\
&  SBR(ours) &\textbf{6.68$\pm$0.53} & \textbf{5.71$\pm$0.44} & \textbf{4.68$\pm$0.51} & \textbf{2.56$\pm$0.34} \\
\hline

\noalign{\medskip}
\multirow{3}{*}{Flowers}
&  L2-SP      &-1.10$\pm$1.02 & 0.91$\pm$0.09 & 0.09$\pm$0.33 & 0.31$\pm$0.07 \\
&  DELTA     &-0.26$\pm$0.94 & -0.19$\pm$0.18 & -0.06$\pm$0.45 & -0.12$\pm$0.23 \\
&  SBR(ours) &\textbf{1.72$\pm$1.08} & \textbf{2.56$\pm$0.66} & \textbf{1.88$\pm$0.18} & \textbf{0.73$\pm$0.18} \\
\hline


\end{tabular}
\end{center}
\end{table}

\subsection{From ImageNet to the Target Task}
Firstly, we used ImageNet-1k \cite{DBLP:journals/ijcv/RussakovskyDSKS15/imagenet} pre-trained model as the source model. ImageNet is widely used large-scale dataset with natural images. Table \ref{table:main_exp_imagenet} shows the increment of various methods compared to the accuracy of fine-tuning with L2 regularization. Weight decay for L2 regularization is 0.0001. The base learning rate for Stanford Dogs \cite{dogs} is 0.001, 0.01 for CUB200 \cite{cub200} and FGVC Aircraft \cite{aircraft}, and 0.1 for Stanford Cars \cite{cars} and Oxford Flowers \cite{flowers}.

The performance gains of methods which used the source model as a reference for the regularization (L2-SP\cite{l2sp}, DELTA\cite{delta}) without considering negative transfer\cite{bss} are not significant. In a few experiments, their results are even worse than the baseline.

Even though the best combinations of BSS and other regularization methods are different according to datasets and sampling rate, BSS performs well on various dataset with the best BSS configuration. Compared to BSS, our method significantly outperform for all configurations. Like BSS, our method was able to get more performance gain in small sampling rate. This is because SBR is more effective in preventing overfitting even with small samples.

\subsection{Using Different Source Model}
\begin{figure}[!t]
  \centering
  \includegraphics[width=0.99\linewidth]{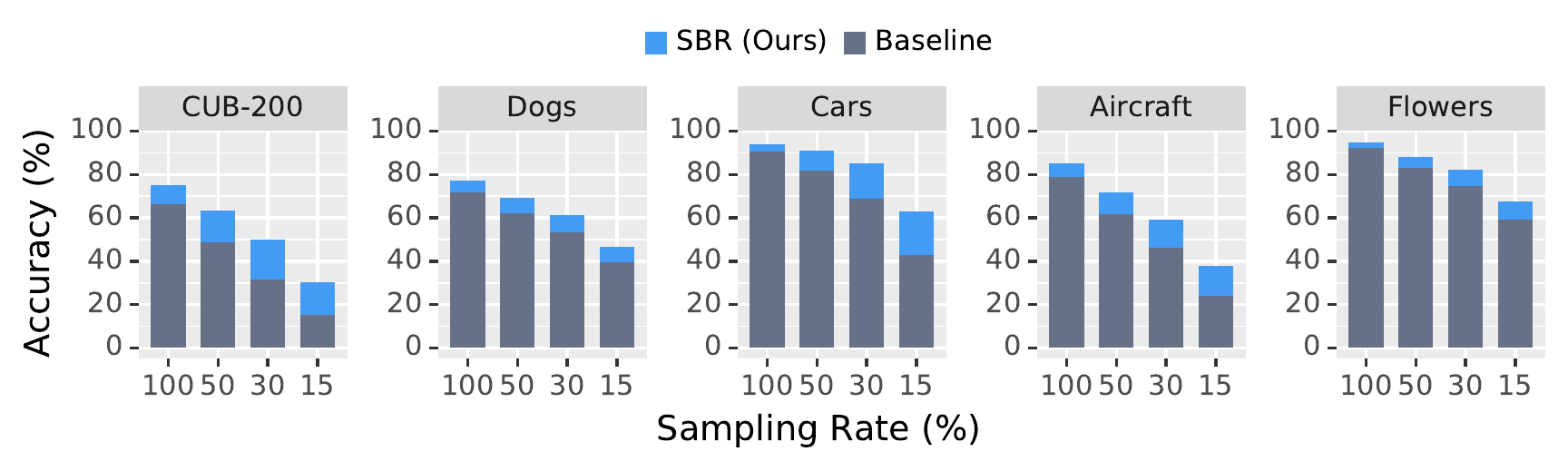}
\caption{
  The performance of the baselines and improvements by applying SBR. The source model is pre-trained on Places365. Our method consistently performs better for all configurations.
  }
\label{fig:places365}
\end{figure}

To show the generality of our method regardless of the choice of the source model, we changed the source model from ImageNet to Places365 \cite{places}. Places365 is a large-scale dataset for predicting the place of the image rather than object classes. As the domain gap between the source and the target is larger than ImageNet, the base learning rate for all target task is set to 0.1. Figure \ref{fig:places365} shows results of this experiment. Even when using a different source model, our method consistently performs better than the baseline.

\subsection{Further Exploration}

We investigate the effects on the dataset size and various similarity measures in more detail here. 
Note that ResNet-18 trained on ImageNet has been used as the source model and our method is tested on CUB-200-2011 dataset in this subsection.

\subsubsection{The regularization strength vs. the size of dataset} \label{sec:beta vs samples}
\begin{figure}[!t]
  \centering
  \includegraphics[width=0.95\linewidth]{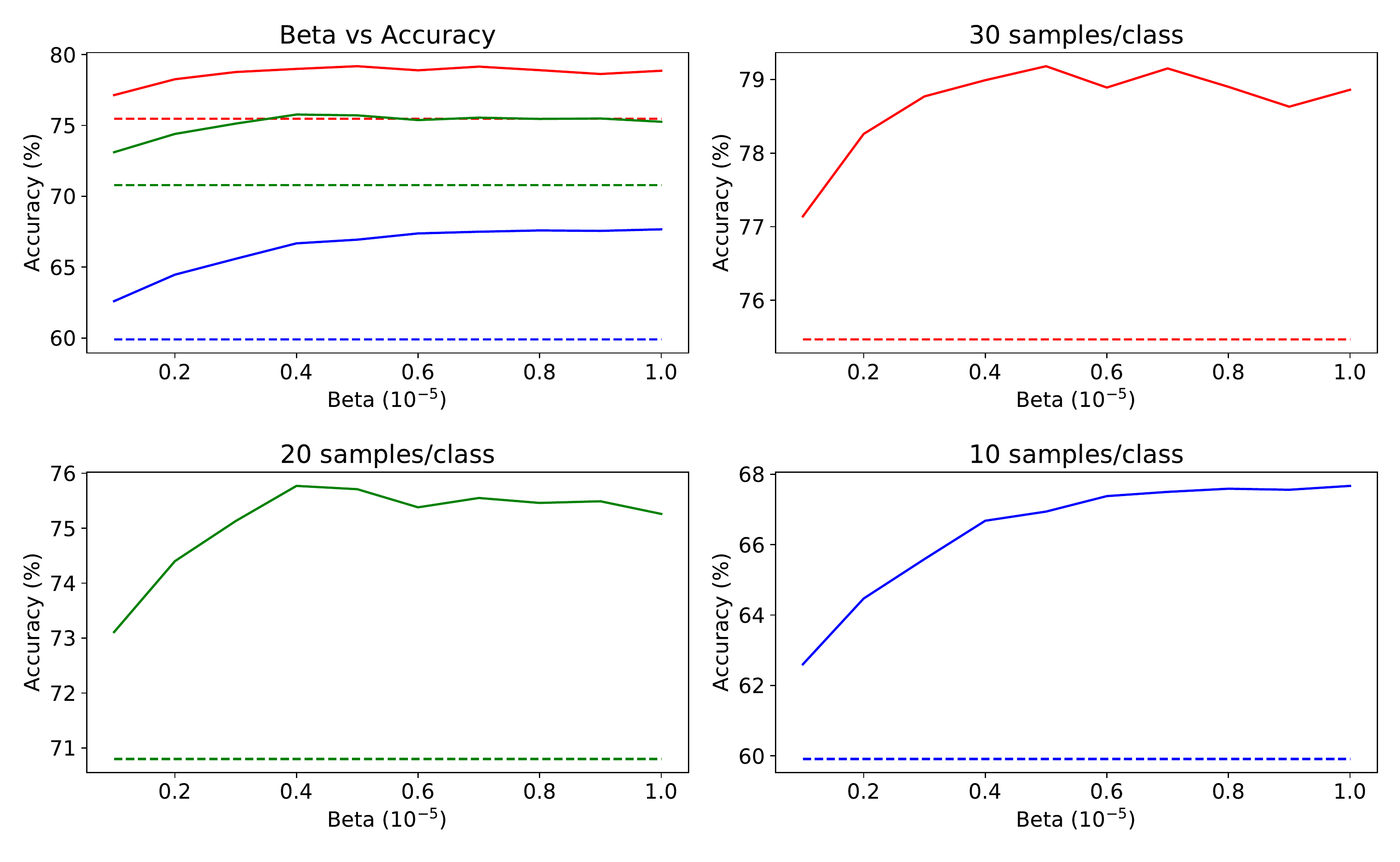}
\caption{
  Effects of the regularization according to dataset size. Red, green and blue line shows the change of accuracy on 30, 20 and 10 samples per class respectively. Dotted lines represent accuracy of the baseline. The left-top graph shows all three experiments and remains show the graph for each experiment in a fine-scale.
  }
\label{fig:reg-vs-size}
\end{figure}

It is believed that learning with a small amount of data is likely to overfit while resulting in low test performance. 
This means that proper regularization might alleviate this problem and a task with fewer samples can benefit more from this regularization.
We investigate the effects of the regularization strength $\beta$ for various dataset sizes to figure out how our method behaves.
In Fig. \ref{fig:reg-vs-size}, we can see that learning with a smaller dataset is more sensitive to the regularization strength $\beta$ while one with a larger dataset shows a flat response relatively over a wide range of $\beta$.
Another observation from this result is that our method leads to larger improvement for learning with a smaller dataset over the baseline fine-tuned without the proposed regularization (represented as dotted lines in Fig. \ref{fig:reg-vs-size}). 
It implies that our method can be a useful way to improve learning with a small amount of data and the regularization strength should be chosen carefully to exploit the full potential of our method.

\begin{table}
\begin{center}
\caption{Performance according to similarity metrics (\textit{N}: samples/class)}
\label{table:exp_diff_similarity}
\begin{tabular}{cccc}
\hline\noalign{\smallskip}
\multirow{2}{*}{Metric} & \multicolumn{3}{c}{Test accuracy (\%)}\\
\cmidrule(l){2-4} & \textit{N}=10 & \textit{N}=20 & \textit{N}=30 \\
\noalign{\smallskip}
\hline
\noalign{\smallskip}
Squared Euclidean & \quad\textbf{67.67$\pm$0.47} & \quad\textbf{75.77$\pm$0.11} & \quad\textbf{79.18$\pm$0.33} \\
Cosine similarity & \quad67.03$\pm$0.88 & \quad75.55$\pm$0.29 & \quad78.81$\pm$0.27 \\
Inner product     & \quad60.25$\pm$0.48 & \quad71.22$\pm$0.11 & \quad75.51$\pm$0.16 \\
\hline
Baseline          & \quad59.65$\pm$0.86 & \quad70.98$\pm$0.33 & \quad75.47$\pm$0.27 \\
\hline
\end{tabular}
\end{center}
\end{table}

\subsubsection{Using different similarity metrics} \label{sec:exp_diff_similarity}
There are many different ways to measure the pairwise similarity. 
We tested cosine similarity and inner product as the similarity metrics by replacing the squared Euclidean distance we have used so far.
Note that the cosine similarity and inner product values should be used with a negative sign so that the regularization term can penalize dissimilar pairs.
Table \ref{table:exp_diff_similarity} shows the test accuracy values after proper scaling of the regularization strength for each similarity metric.
Our method with the cosine similarity performs comparably to one with the squared Euclidean distance.
Using the inner product leads to slightly better performance than the baseline fine-tuned without the proposed regularization, but it is far behind the squared Euclidean and the cosine similarity. 


%% file: 05_discussions.tex
\section{Discussions}

\subsection{SBR with Squared Euclidean Distance} \label{sec:SBR with L2}
In table \ref{table:exp_diff_similarity}, we showed that using squared Euclidean distance as a dissimilarity measure performs better than other measures. Beyond this empirical effectiveness, using the L2 norm is practically useful as we can convert it to a simpler form. SBR loss with squared Euclidean distance is formulated as follows:

\begin{align}
\label{eq:SBR loss with L2}
  L_{sbr}^{l2} = \sum_{c} {1\over N^{pair}_c }\sum_{(x_i,x_j) \in B_c} {1\over 2}||f(x_i, \textbf{w}_f) - f(x_j,  \textbf{w}_f)||^2 
\end{align},
and when it is differentiated with respect to a feature $f_i = f(x_i, \textbf{w}_f)$ where $y_i = c$,

\begin{align}
{\partial L_{sbr}^{l2} \over \partial f_i} &= {1 \over N^{pair}_c} \cdot 2((N_c-1)\cdot f_i - \sum_{x_j \in B_c \setminus \{x_i\}} f_j) \\ &= {2 \over N_{c}(N_c -1 )} (N_c\cdot f_i - \sum_{x_j \in B_c} f_j) \\
&= {2 \over {N_c - 1}} (f_i - {1 \over N_c} \sum_{x_j \in B_c} f_j)\\
&= {2 \over {N_c - 1}} (f_i - C_{c}), 
\end{align}
where $C_{c}$ is the average of the sample features (i.e., $C_{c} = {1 \over N_c} \sum_{x_i \in B_c} f(x_i , \textbf{w}_f)$). This is the center point of the features which have the same class within a batch. Therefore, we can simply rewrite the $L_{sbr}^{l2}$ using $C_c$ as follows:
\begin{align}
\label{eq:SBR with l2 and center}
L_{sbr}^{l2} = \sum_{c} {1 \over {N_c - 1}} \sum_{x_i \in B_c} ||f(x_i, \textbf{w}_f) - C_{c}||^2.
\end{align}

This shows that SBR using the L2 norm can be converted to the L2 regularization with the center of the features for each class and  normalization factor ${1 \over {N_c - 1}}$. While the original SBR with squared Euclidean distance in Eq. (\ref{eq:SBR loss with L2}) required O($N_c^2$) calculations for the pairwise relationship, the computational cost for Eq. (\ref{eq:SBR with l2 and center}) reduced to $O(N_c)$. This formulation is similar to the center loss in \cite{DBLP:conf/eccv/WenZL016/center-loss} for a face recognition task. However, our method is different from theirs as we consider the samples only within a single batch for the centers rather than maintaining class centroids throughout the entire training.


\begin{table}
\begin{center}
\caption{Comparison between reducing learning rate and gradient in feature extractor. Reducing gradient is more effective than reducing learning rate when SBR is applied.}
\label{table:fe_gr_comparison}
\begin{tabular}{C{0.2\textwidth}|R{0.08\textwidth}R{0.08\textwidth}R{0.12\textwidth}R{0.2\textwidth}}

\hline\noalign{\smallskip}
samples/class & $\kappa$ & $\alpha$ & $\beta$ & top1(\%)\\
\noalign{\smallskip}
\hline
\noalign{\smallskip}
\multirow{4}{*}{30}
      &  0.1     &   1       &   0       &   {75.47$\pm$0.27} \\
      &  1       &   0.1     &   0       &   {75.66$\pm$0.16}\\
      &  0.1     &   1       &   0.001      &  {77.70$\pm$0.19}\\
      &  1       &   0.1     &   0.0001     &  {\textbf{78.86$\pm$0.14}}\\
\hline
\multirow{4}{*}{20}
      &  0.1     &   1       &   0       &   {70.80$\pm$0.33} \\
      &  1       &   0.1     &   0       &	{71.38$\pm$0.2} \\
      &  0.1     &   1       &   0.001      &  {74.47$\pm$0.21}\\
      &  1       &   0.1     &   0.0001     &  {\textbf{75.26$\pm$0.21}}\\

\hline
\multirow{4}{*}{10}
      &  0.1     &   1       &   0       &   {59.91$\pm$0.86} \\
      &  1       &   0.1     &   0       &	 {60.19$\pm$0.17} \\
      &  0.1     &   1       &   0.001      &	65.77$\pm$0.24 \\
      &  1       &   0.1     &   0.0001     &	\textbf{67.67$\pm$0.47}  \\
\hline
\end{tabular}
\end{center}
\end{table}

\subsection{The Effect of the Reducing Gradient} \label{sec:diff_gr_fe}
We proposed to reduce gradient from the classifier to feature extractor by a factor of $\alpha$ (which was set to 0.1 in every experiment except for what we mentioned) for increasing the generalization performance of feature extractor. We also used 0.1 times smaller learning rate on the feature extractor compared to the classifier. This is the general configuration for transfer learning as the parameters of classifier are randomly initialized. We empirically found that this configuration works well in most cases. If the gradient reduction with $\alpha$ and the reduced learning rate for feature extractor at the same time, the gradient to feature extractor diminishes drastically. Therefore, when the gradient reduces is applied, we did not decrease the learning rate of feature extractor to match the same amount of gradient from the classifier compared to baseline.

When SGD without momentum is used as the optimizer for our algorithm, above explanation can be formulated using parameter update rules:

\begin{align}
\label{eq:param_update_rule}
  \textbf{w}_g' &= \textbf{w}_g + \eta_g  \cdot (\nabla L_{cls} + \lambda_g \nabla \Omega), \\ 
  \textbf{w}_f' &= \textbf{w}_f + \eta_f  \cdot (\alpha\nabla L_{cls} + \beta\nabla L_{sbr} + \lambda_f  \nabla \Omega), \label{eq:param_update_rule_fe}
\end{align}
where $\eta_g$ and $\eta_f$ are learning rate for the classifier and the feature extractor, respectively. In a same manner, $\lambda_{\cdot}$ is a weight decay for each part. If we reduce the learning rate of the feature extractor by factor of $\kappa$ (i.e., $\eta_f = \eta_g / \kappa$), Eq. (\ref{eq:param_update_rule_fe}) can be reformulated as follows: 

\begin{align}
  \textbf{w}_f' &= \textbf{w}_f + {\eta_g
   \over  \kappa}  \cdot (\alpha\nabla L_{cls} + \beta \nabla L_{sbr} + \lambda_f  \nabla \Omega) \\
  &= \textbf{w}_f + \eta_g  \cdot ({\alpha \over  \kappa} \nabla L_{cls} + {\beta \over  \kappa} \nabla L_{sbr} + {\lambda_f \over  \kappa}  \nabla \Omega) \\
  &= \textbf{w}_f + \eta_g  \cdot ({\hat{\alpha}} \nabla L_{cls} + {\hat{\beta}} \nabla L_{sbr} + {\hat{\lambda_f}}  \nabla \Omega)
\end{align}
where $\hat{\alpha}$, $\hat{\beta}$ and $\hat{\lambda_f}$ are rescaled parameter reduced by the ratio of $\kappa$. This means that the hyper-parameter set $(\eta _g / \kappa, \alpha, \beta, \lambda _f)$ is equivalent to $(\eta _g, \alpha / \kappa, \beta / \kappa, \lambda _f / \kappa)$. Generally, as the same weight decay is applied for all layers (i.e., $\lambda _g = \lambda _f$), reducing gradient is different from reducing learning rate. 

For comparing the effect of reducing the gradient and the learning rate for the feature extractor, we performed additional experiments with equivalent hyper-parameters without controlling weight decay. Experiments are performed with CUB-200 dataset using ImageNet pre-trained ResNet-18. As shown in Table \ref{table:fe_gr_comparison}, the accuracy of reducing gradient without SBR is similar to that of reducing the learning rate. With SBR, however, reducing gradient boosts accuracy, which implies that reducing the influence of cross-entropy loss to the feature extractor and training with SBR helps the feature extractor to learn more general features.

%% file: 06_conclusions.tex
\section{Conclusions}
In this paper, we proposed a simple but effective regularization method for transfer learning. Our method exploited the pairwise relation of target samples rather than relying on the source model parameters or activations. We defined a loss term to encourage the pairwise similarity between the same class samples in the feature space of the transferred model. This helped the fine-tuning to achieve better generalization without increasing the risk of potential negative transfer from the source model. Experiments showed that the proposed SBR outperformed other methods and was effective particularly for learning with a small number of data. Beyond the fine-tuning, the standard supervised learning or the semi-supervised learning might benefit from the proposed regularization, which can be promising future research directions.

%% file: 10_supplement.tex
%






%
\title{Appendix}
\titlerunning{Appendix}

\author{}
\institute{}

\maketitle

\begin{figure}[b!]
\caption*{Sampling rate 15\%}
\vspace{-0.3cm}
\begin{subfigure}{.195\textwidth}
  \includegraphics[width=\linewidth]{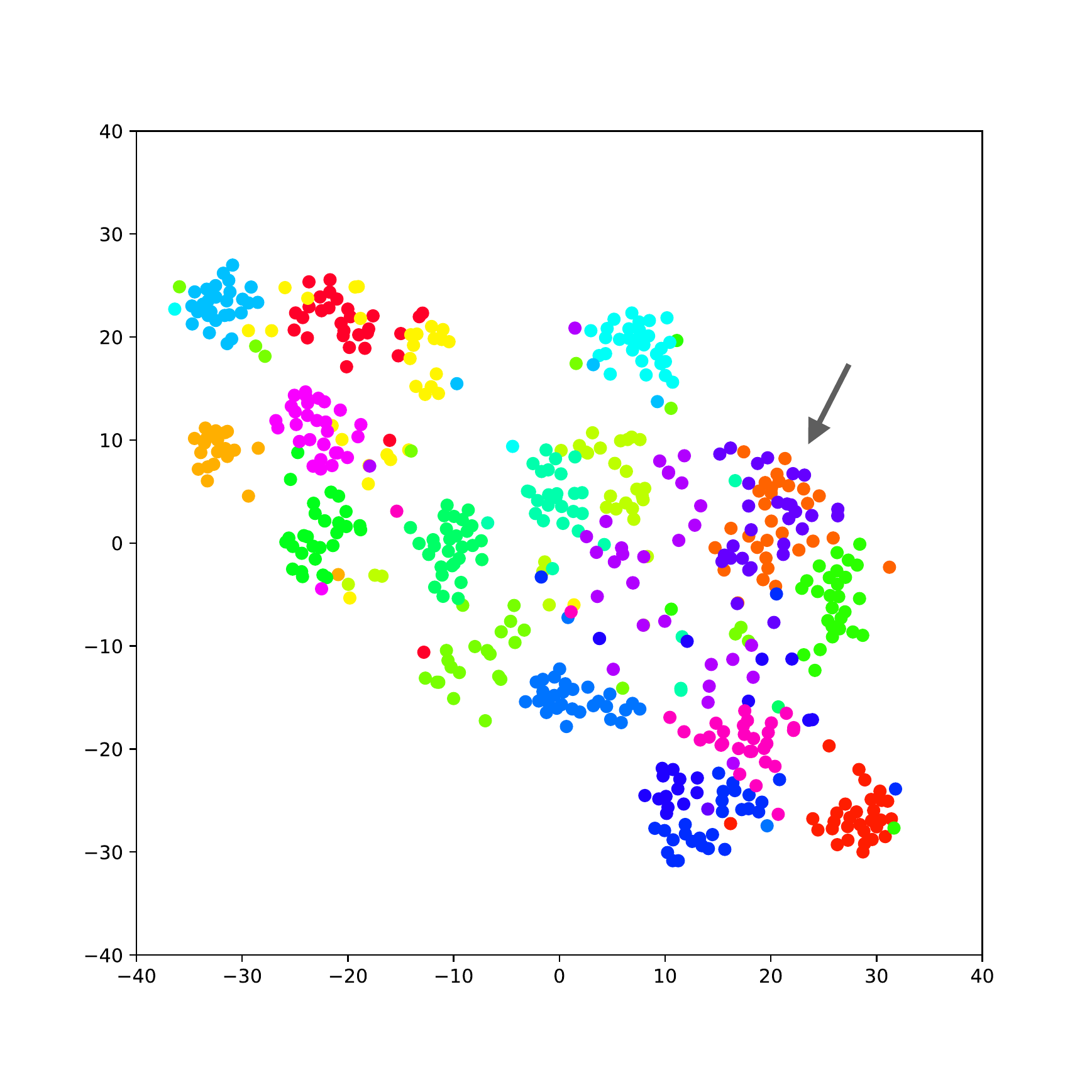}
  \caption{Baseline (L2)}
\end{subfigure}
\begin{subfigure}{.195\textwidth}
  \includegraphics[width=\linewidth]{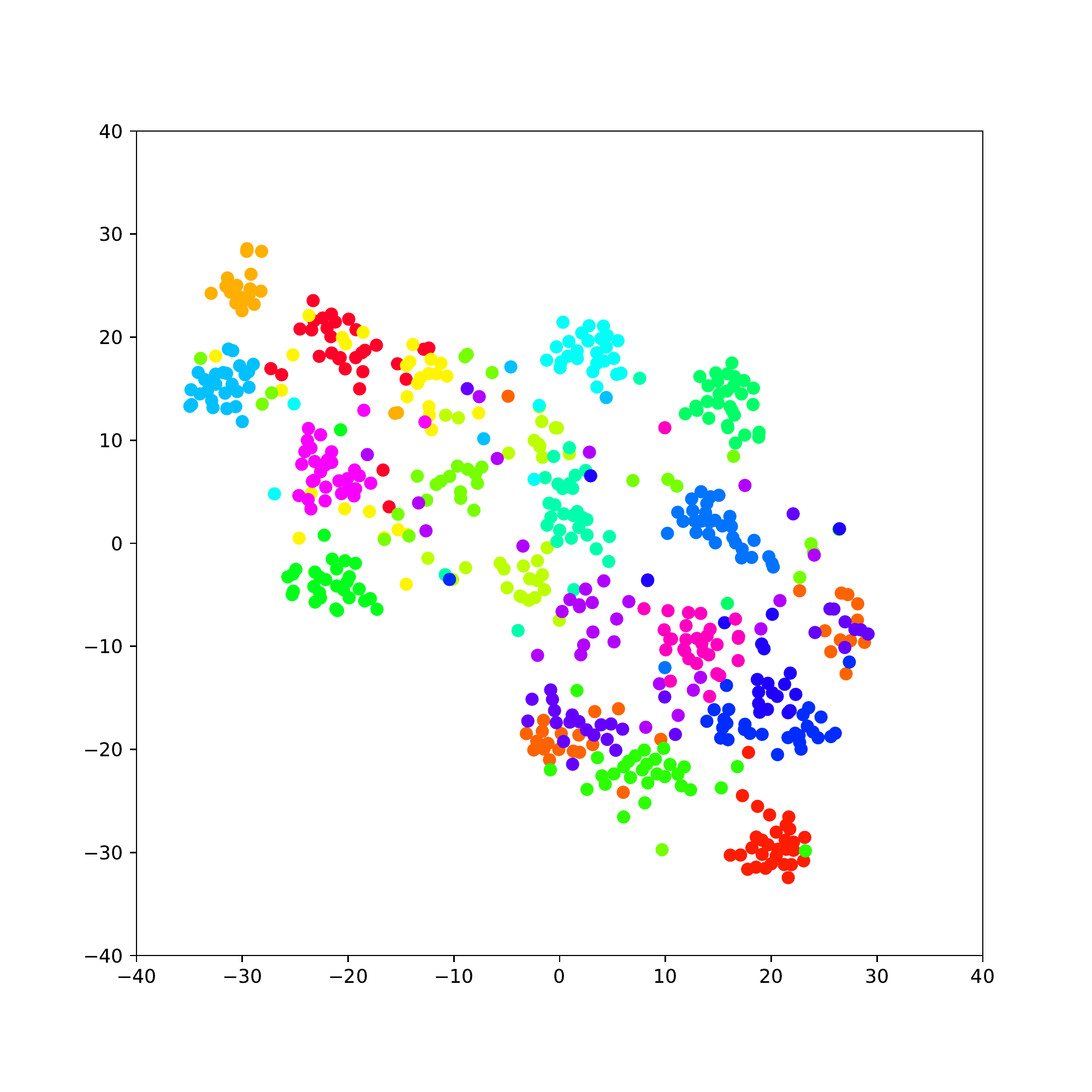}
  \caption{L2-SP}
\end{subfigure}
\begin{subfigure}{.195\textwidth}
  \includegraphics[width=\linewidth]{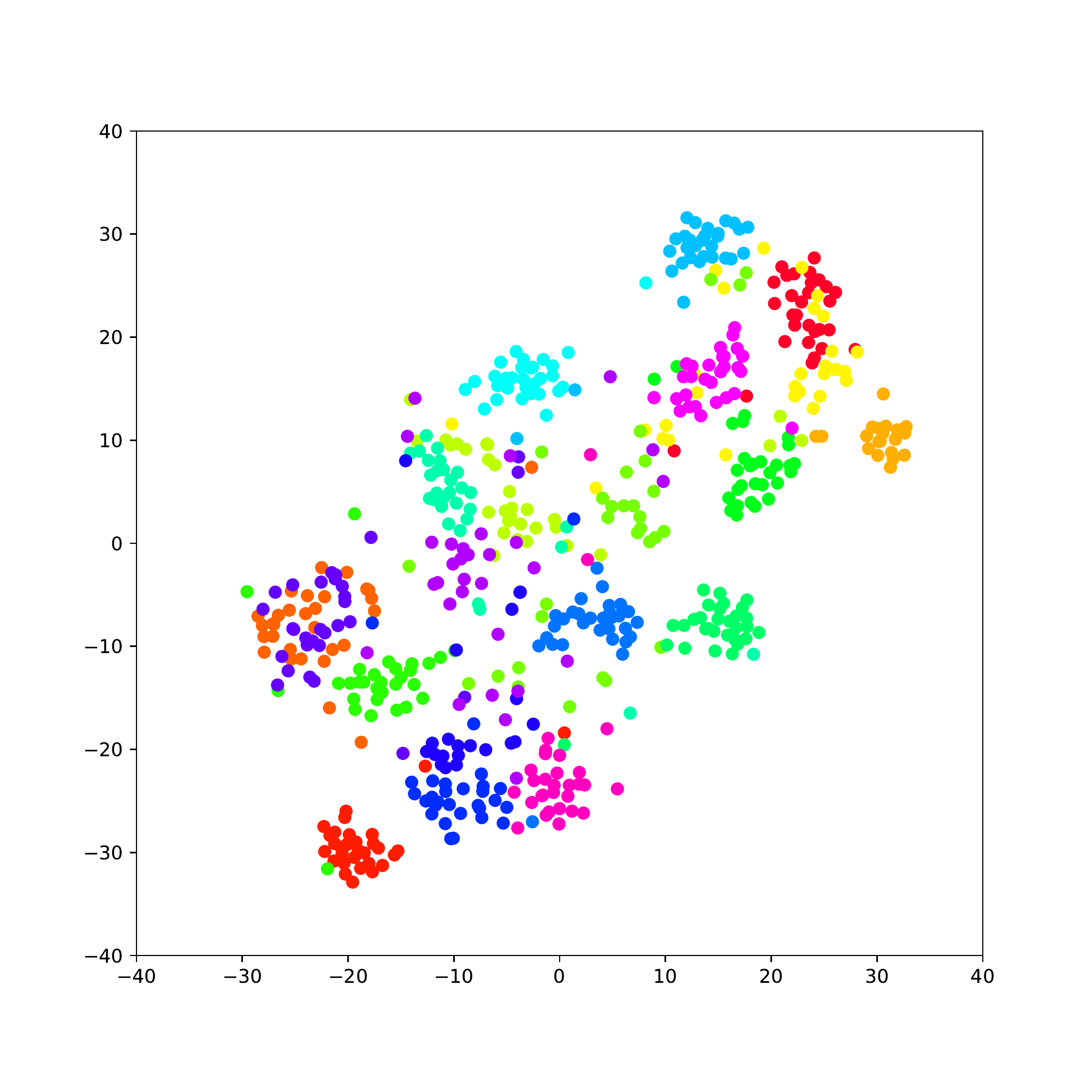}
  \caption{DELTA}
\end{subfigure}
\begin{subfigure}{.195\textwidth}
  \includegraphics[width=\linewidth]{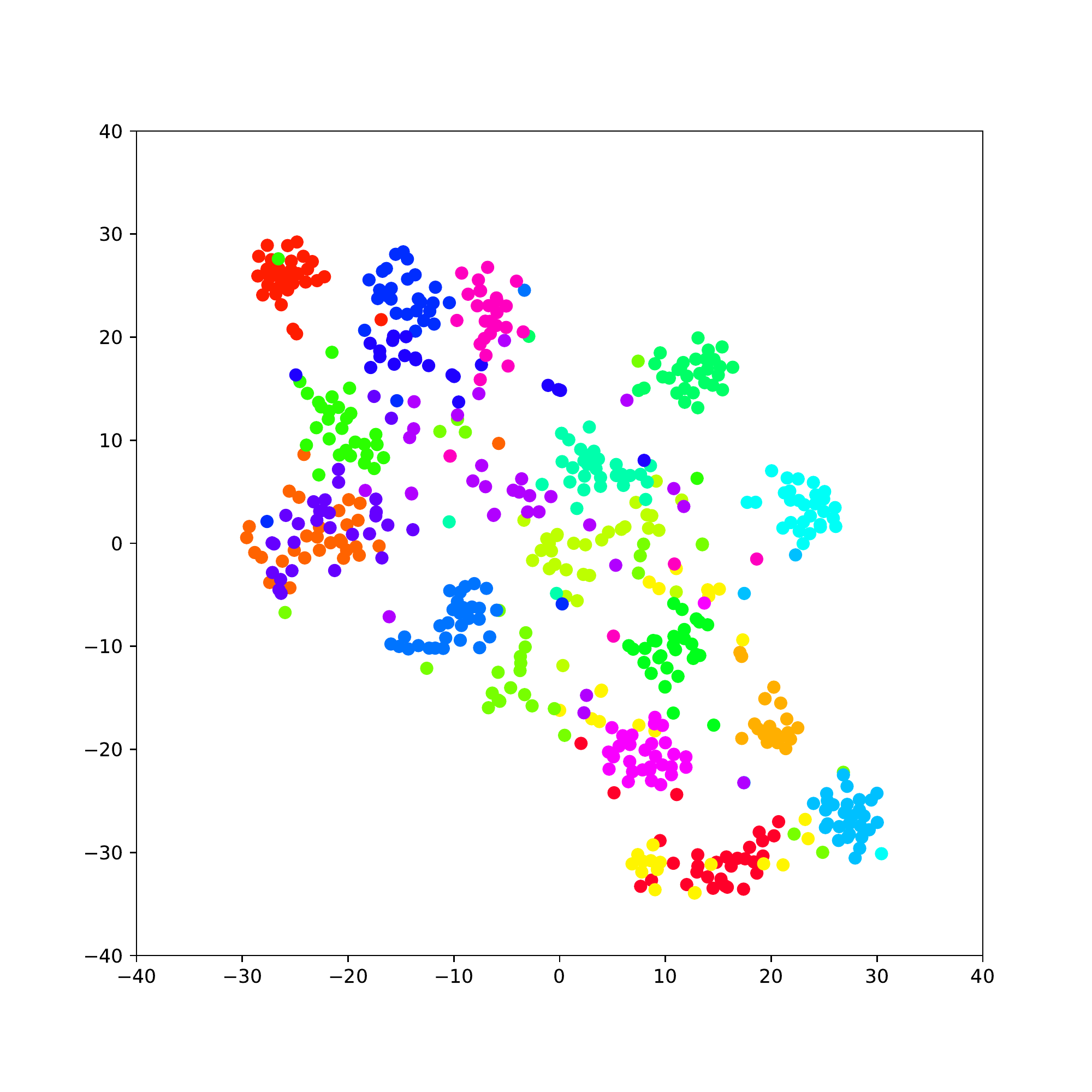}
  \caption{L2+BSS}
\end{subfigure}
\begin{subfigure}{.195\textwidth}
  \includegraphics[width=\linewidth]{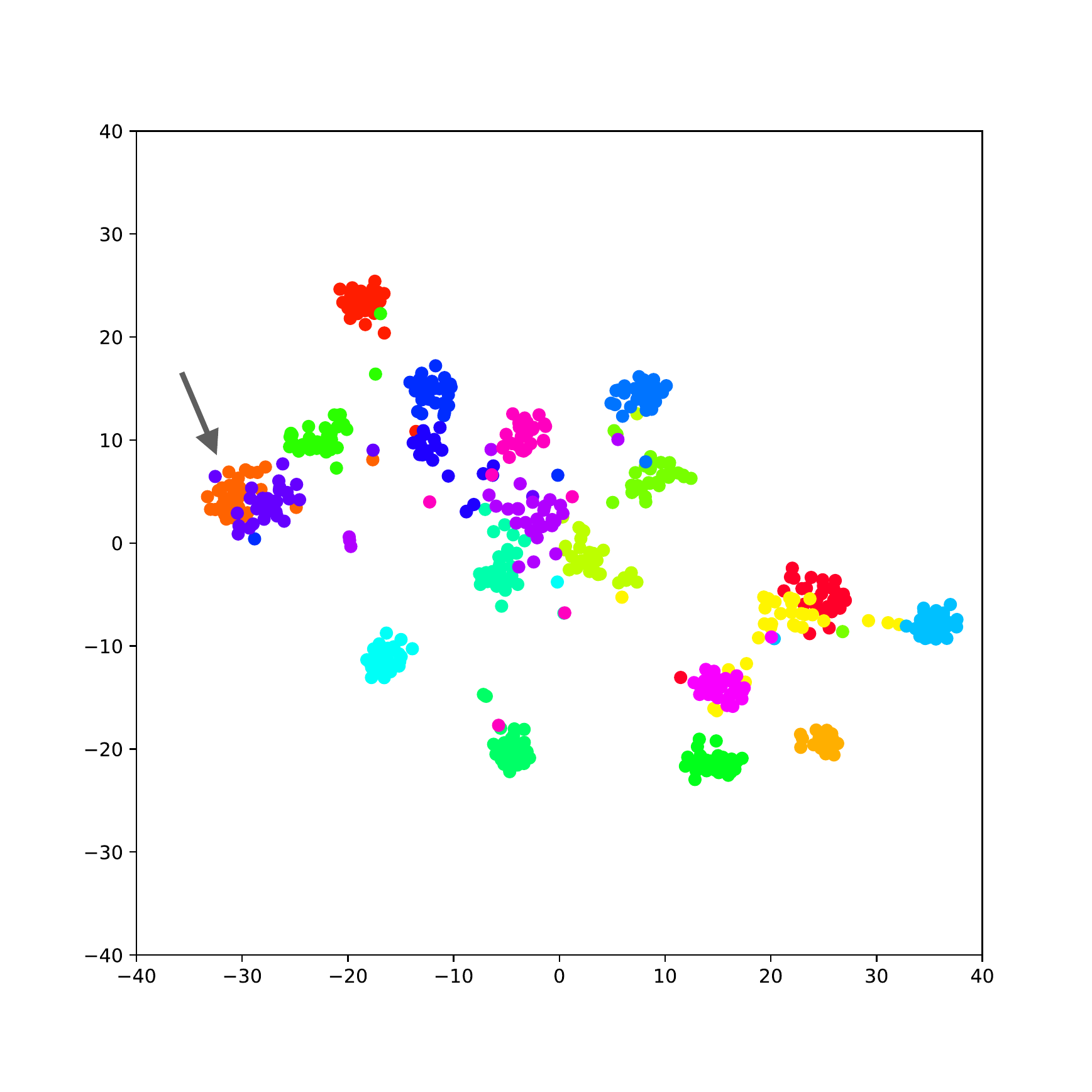}
  \caption{SBR (ours)}
\end{subfigure}
\vspace{0.2cm}
\caption*{Sampling rate 100\%}
\vspace{-0.3cm}
\begin{subfigure}{.195\textwidth}
  \includegraphics[width=\linewidth]{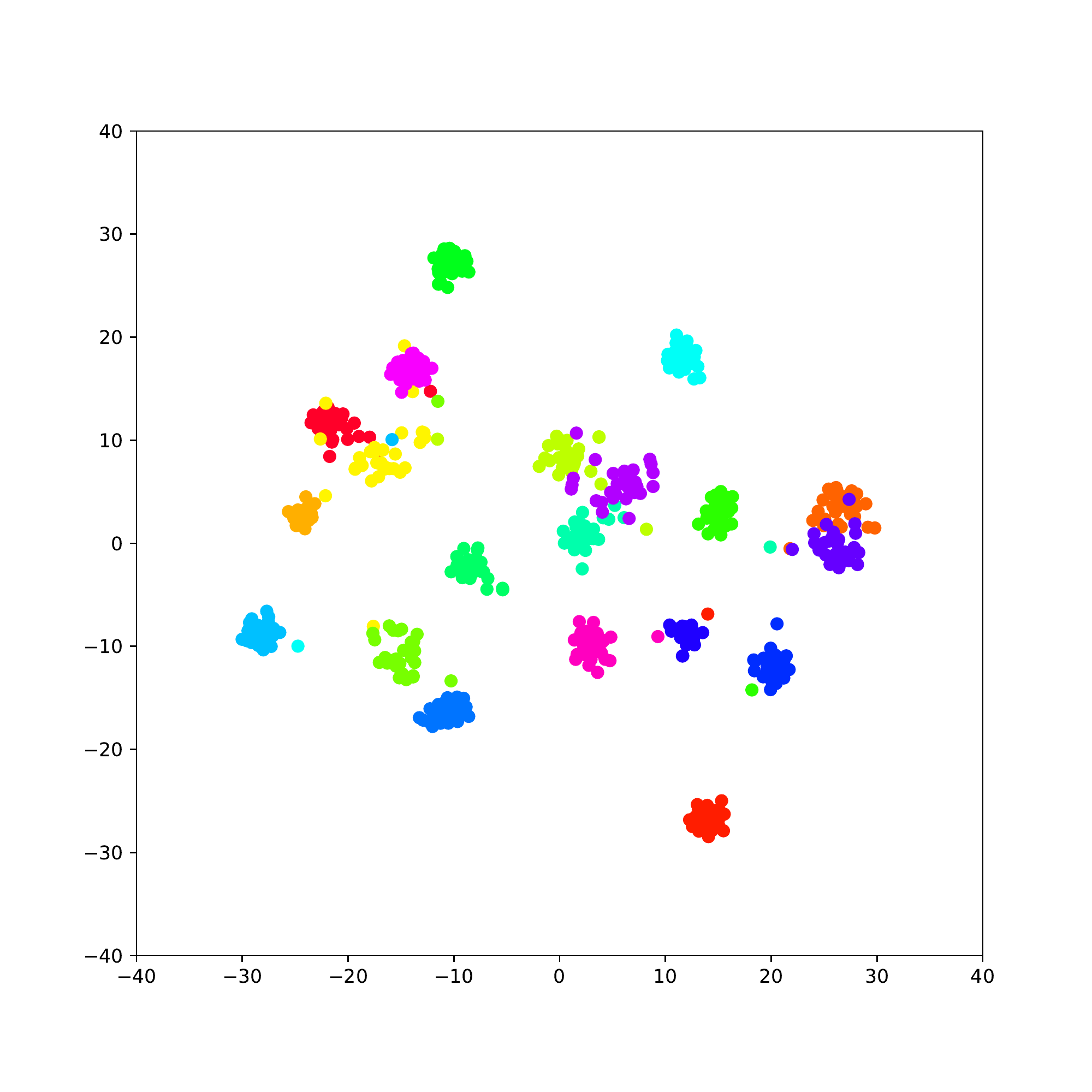}
  \caption{Baseline (L2)}
\end{subfigure}
\begin{subfigure}{.195\textwidth}
  \includegraphics[width=\linewidth]{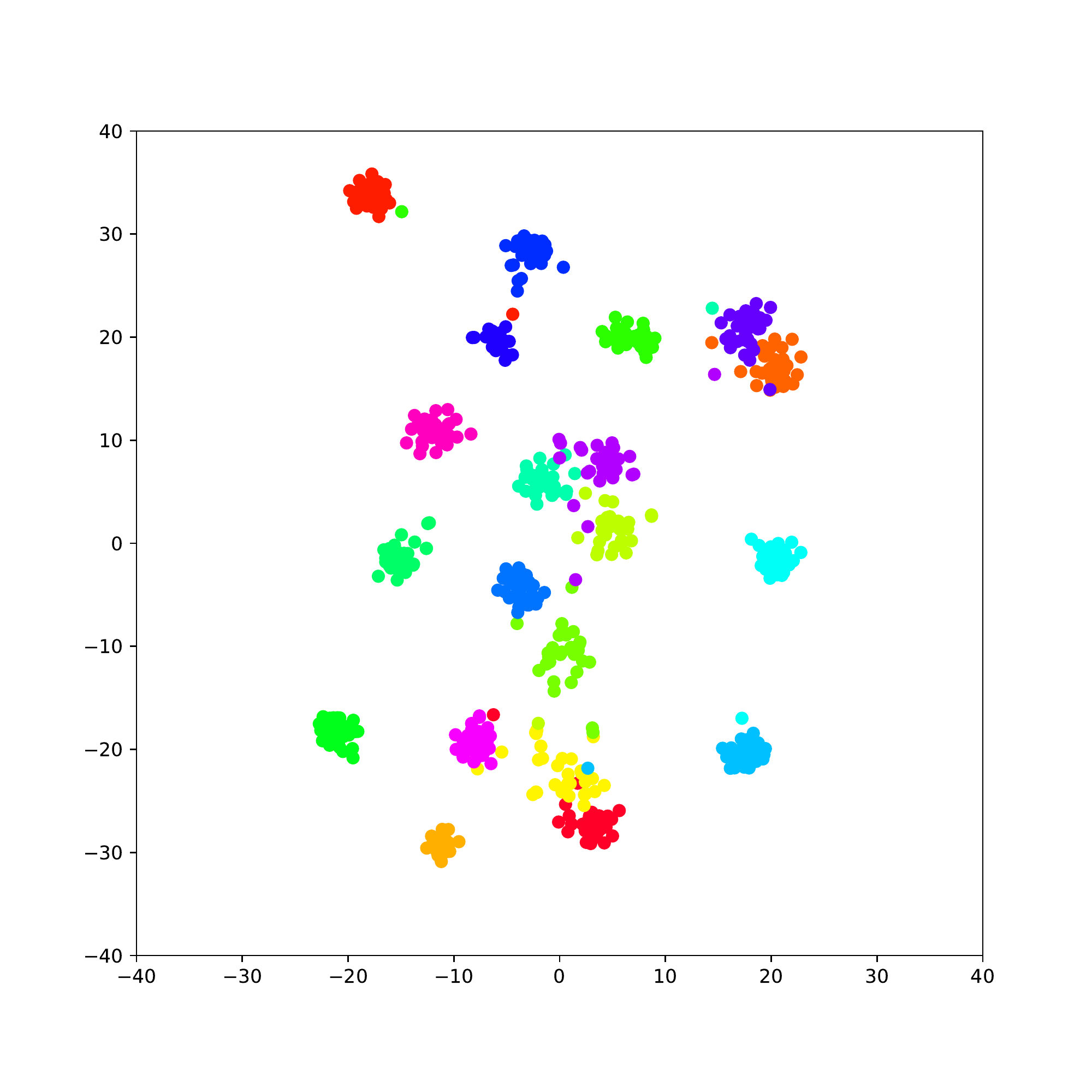}
  \caption{L2-SP}
\end{subfigure}
\begin{subfigure}{.195\textwidth}
  \includegraphics[width=\linewidth]{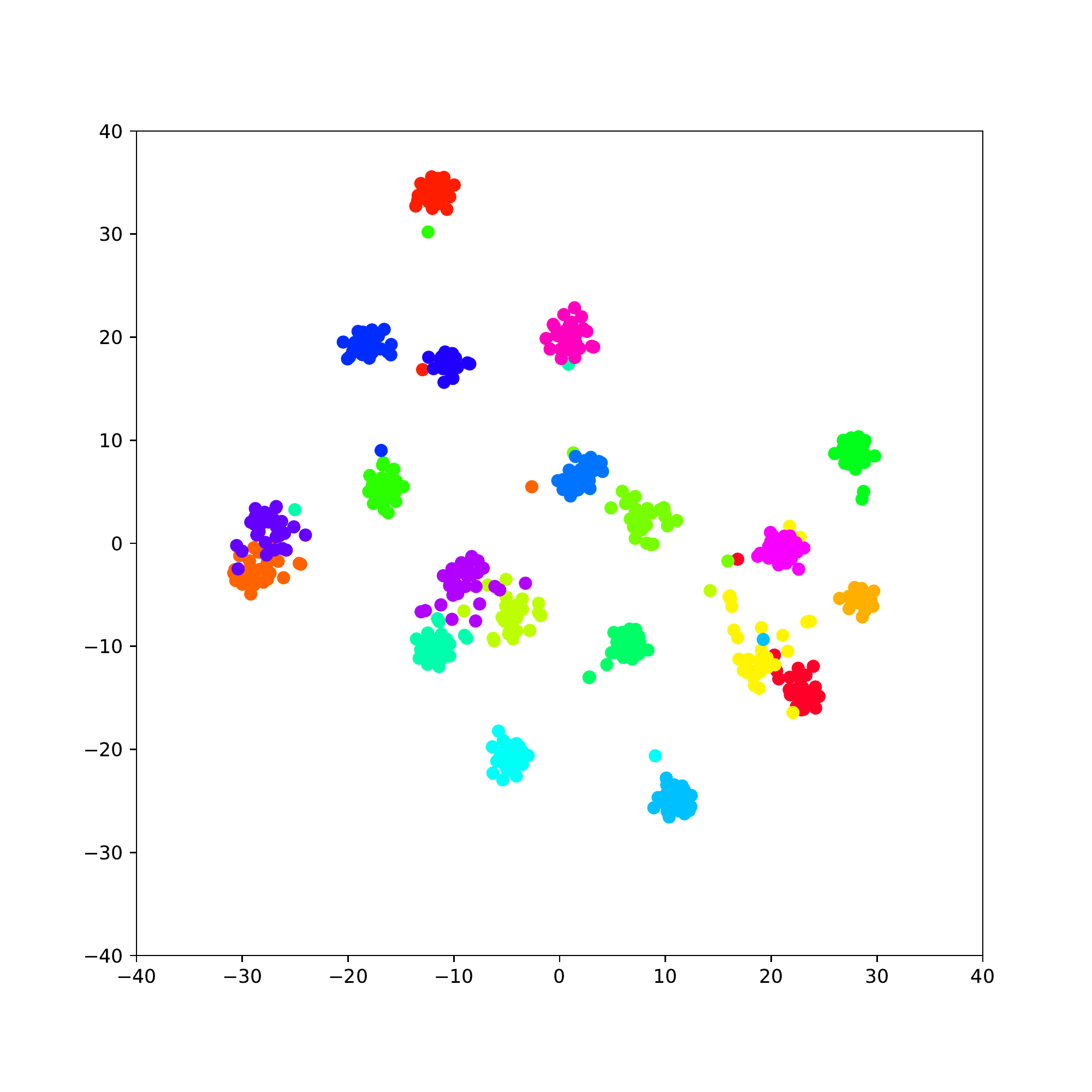}
  \caption{DELTA}
\end{subfigure}
\begin{subfigure}{.195\textwidth}
  \includegraphics[width=\linewidth]{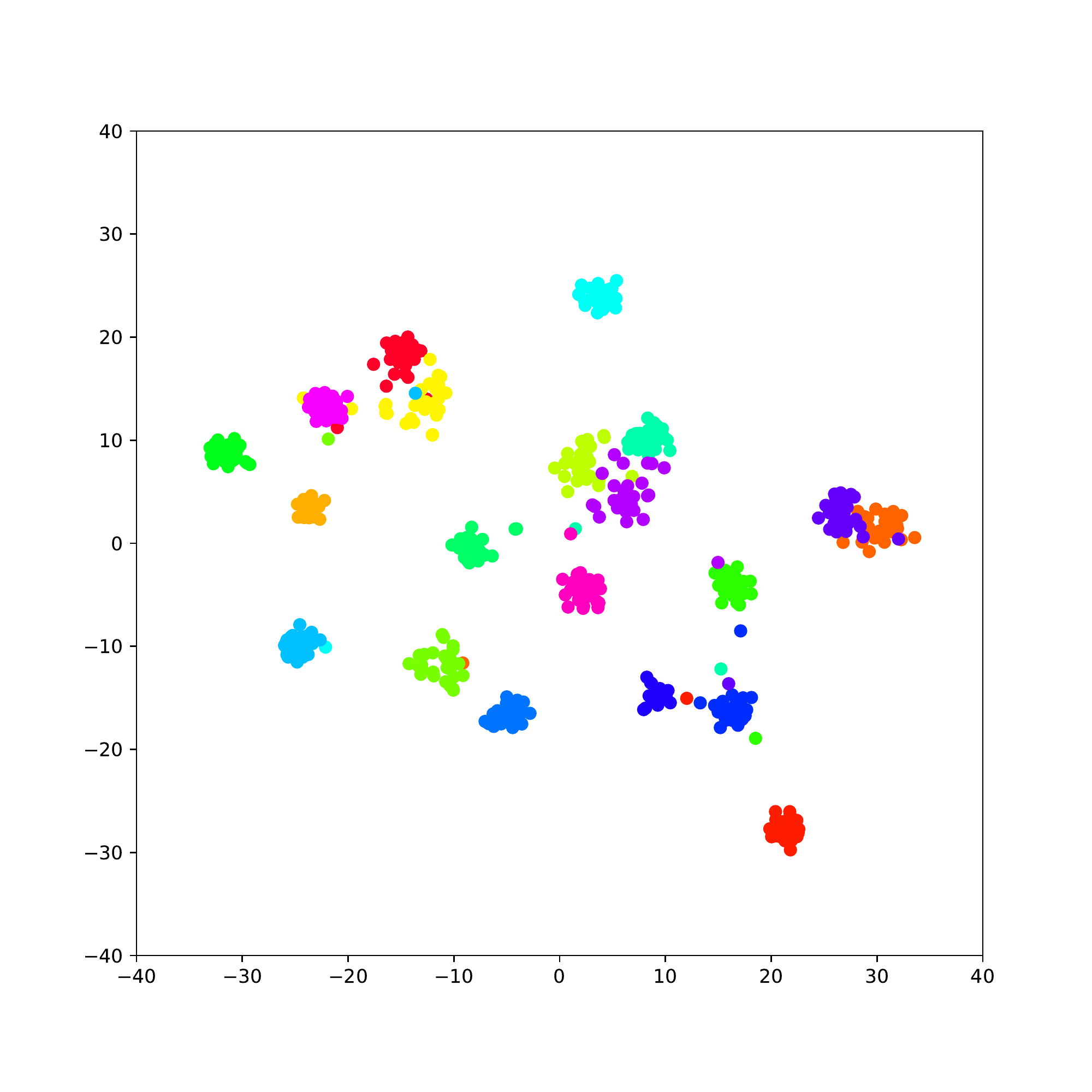}
  \caption{L2+BSS}
\end{subfigure}
\begin{subfigure}{.195\textwidth}
  \includegraphics[width=\linewidth]{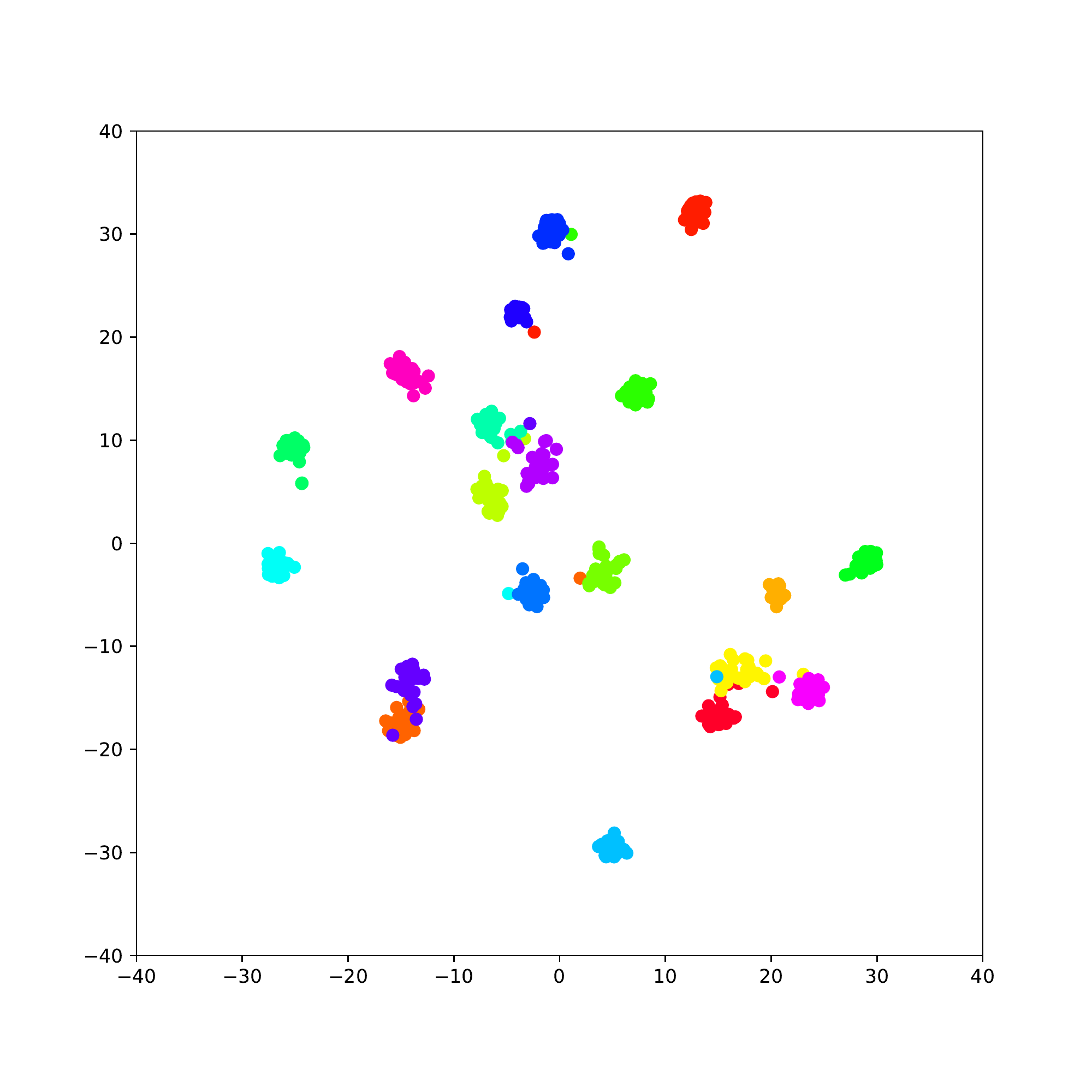}
  \caption{SBR (ours)}
\end{subfigure}
\caption{Visualization of features. The features are drawn using test images of CUB200 and 20-classes are evenly chosen for visualization (Chosen classes are the same for all figures). Our method effectively reduces the variance of features of test images and improves the generalization performance. Furthermore, some mixed classes are separated after applying SBR (see arrow). Best viewed in color.}
\label{fig:feature_visualization}
\end{figure}

\section{Visualization of features}
In Fig. \ref{fig:feature_visualization}, we visualize outputs of the feature extractor for test images to show the distribution of features depending on various methods. Features are visualized using t-SNE \cite{maaten2008visualizing} with evenly selected 20-classes and these features are obtained from ResNet-50 trained on CUB-200. In the case of the number of training examples is small, our method effectively reduced the variation of samples within the same class, which means the trained model can generalize well on the test set. Furthermore, some mixed classes in naive fine-tuning are separated after applying our method (orange and purple classes (see arrow) in Fig. \ref{fig:feature_visualization}(a) and (e)).
When the number of training samples increases, the baseline can also reduce the variation in clusters and increase the generalization performance as shown in Fig. \ref{fig:feature_visualization}(f) and this is consistent with our intuition. Even in the case that 100\% of training examples are used, our method still seems to be helpful to shrink the class boundary.



\section{Baseline Fine-tuning Results}
As the base learning rate differs depending on the source and target task \cite{DBLP:conf/iclr/Li20/rethinking-ft}, we conducted experiments to find the best learning rate for each target task with the learning rate in $\{0.1, 0.01, 0.001\}$. For all experiments, the learning rate for the feature extractor is reduced by 0.1 from the base learning rate. The ImageNet and Places365 pretrained models are used as sources. Table \ref{table:lr_exp_imagenet} and \ref{table:lr_exp_places365} show results for ImageNet and Places365, respectively. Although the best learning rate varies depending on the sampling rate, the learning rate which performs well in different sampling rates was chosen for experiments in the main paper. (e.g., 0.001 for Dogs and 0.01 for Aircraft in ImageNet). In Places365 experiments, 0.1 is the best learning rate for all experiments. 

\begin{table}[!h]
\begin{center}
\caption{Test accuracy (\%) of fine-tuning using ImageNet as a source according to the learning rate}
\label{table:lr_exp_imagenet}
\begin{tabular}{C{1.5cm}C{1.5cm}R{2cm}R{2cm}R{2cm}R{2cm}}
\hline\noalign{\smallskip}
\multirow{2}{*}{Dataset} & \multirow{2}{*}{\thead{Learning \\ Rate}} & \multicolumn{4}{c}{Sampling Rate}\\
            \cmidrule(l){3-6}    &  & \multicolumn{1}{c}{15\%} & \multicolumn{1}{c}{30\%} & \multicolumn{1}{c}{50\%} & \multicolumn{1}{c}{100\%} \\
\noalign{\smallskip}
\hline
\hline
\noalign{\smallskip}
\multirow{3}{*}{CUB-200}   
&   0.1 & 40.38 $\pm$  0.37 & 56.22 $\pm$  0.22 & 65.37 $\pm$  0.37 & 75.93 $\pm$  0.35 \\
&  0.01 & \textbf{46.42} $\pm$  0.23 & \textbf{62.72} $\pm$  0.12 & \textbf{72.09} $\pm$  0.37 & \textbf{80.13} $\pm$  0.23 \\
& 0.001 & 18.40 $\pm$  1.30 & 35.21 $\pm$  0.89 & 51.44 $\pm$  0.47 & 68.31 $\pm$  0.24 \\
\hline

\noalign{\smallskip}
\multirow{3}{*}{Dogs}   
&   0.1 & 64.44 $\pm$  0.21 & 67.10 $\pm$  0.28 & 70.60 $\pm$  0.60 & 75.68 $\pm$  0.18 \\
&  0.01 & \textbf{81.11} $\pm$  0.40 & 82.62 $\pm$  0.15 & 83.67 $\pm$  0.17 & 84.39 $\pm$  0.07 \\
& 0.001 & 79.51 $\pm$  0.54 & \textbf{84.10} $\pm$  0.34 & \textbf{86.06} $\pm$  0.13 & \textbf{87.60} $\pm$  0.11 \\
\hline

\noalign{\smallskip}
\multirow{3}{*}{Cars}  
&   0.1 & \textbf{36.58} $\pm$  1.07 & \textbf{64.30} $\pm$  0.88 & \textbf{78.76} $\pm$  0.05 & \textbf{89.45} $\pm$  0.20 \\
&  0.01 & 34.74 $\pm$  0.64 & 62.05 $\pm$  0.43 & 77.45 $\pm$  0.33 & 88.48 $\pm$  0.07 \\
& 0.001 &  6.90 $\pm$  0.34 & 14.97 $\pm$  0.57 & 27.76 $\pm$  0.20 & 57.48 $\pm$  0.85 \\
\hline

\noalign{\smallskip}
\multirow{3}{*}{Aircraft}   
&   0.1 & 33.23 $\pm$  1.35 & 55.69 $\pm$  0.51 & 67.97 $\pm$  0.59 & \textbf{82.72} $\pm$  0.66 \\
&  0.01 & \textbf{37.73} $\pm$  0.45 & \textbf{56.79} $\pm$  0.67 & \textbf{68.85} $\pm$  0.64 & 81.63 $\pm$  0.36 \\
& 0.001 & 14.10 $\pm$  0.68 & 25.12 $\pm$  0.55 & 40.21 $\pm$  0.99 & 58.27 $\pm$  0.88 \\
\hline

\noalign{\smallskip}
\multirow{3}{*}{Flowers}
&   0.1 & \textbf{76.45} $\pm$  0.37 & \textbf{86.06} $\pm$  0.37 & \textbf{90.59} $\pm$  0.19 & \textbf{96.25} $\pm$  0.19 \\
&  0.01 & 72.91 $\pm$  0.86 & 84.48 $\pm$  0.25 & 89.94 $\pm$  0.39 & 95.64 $\pm$  0.18 \\
& 0.001 & 26.24 $\pm$  0.78 & 54.98 $\pm$  1.27 & 67.70 $\pm$  1.14 & 83.20 $\pm$  0.32 \\
\hline

\end{tabular}
\end{center}
\end{table}

\begin{table}[!h]
\begin{center}
\caption{Test accuracy (\%) of fine-tuning using Places365 as a source according to the learning rate}
\label{table:lr_exp_places365}
\begin{tabular}{C{1.5cm}C{1.5cm}R{2cm}R{2cm}R{2cm}R{2cm}}
\hline\noalign{\smallskip}
\multirow{2}{*}{Dataset} & \multirow{2}{*}{\thead{Learning \\ Rate}} & \multicolumn{4}{c}{Sampling Rate}\\
            \cmidrule(l){3-6}    &  & \multicolumn{1}{c}{15\%} & \multicolumn{1}{c}{30\%} & \multicolumn{1}{c}{50\%} & \multicolumn{1}{c}{100\%} \\
\noalign{\smallskip}
\hline
\hline
\noalign{\smallskip}
\multirow{3}{*}{CUB-200}   
&   0.1 & \textbf{15.08} $\pm$  0.16 & \textbf{31.65} $\pm$  0.76 & \textbf{48.56} $\pm$  0.59 & \textbf{66.43} $\pm$  0.35 \\
&  0.01 & 10.39 $\pm$  0.79 & 26.62 $\pm$  0.38 & 44.84 $\pm$  0.47 & 64.57 $\pm$  0.33 \\
& 0.001 &  2.43 $\pm$  0.14 &  4.66 $\pm$  0.30 &  7.25 $\pm$  0.39 & 13.50 $\pm$  0.15 \\
\hline

\noalign{\smallskip}
\multirow{3}{*}{Dogs}   
&   0.1 & 32.13 $\pm$  0.27 & 44.92 $\pm$  0.07 & 55.14 $\pm$  0.38 & \textbf{66.34} $\pm$  0.15 \\
&  0.01 & \textbf{33.59} $\pm$  0.26 & \textbf{47.60} $\pm$  0.22 & \textbf{55.44} $\pm$  0.51 & 65.84 $\pm$  0.34 \\
& 0.001 &  6.44 $\pm$  0.43 & 13.48 $\pm$  0.54 & 21.16 $\pm$  0.38 & 39.34 $\pm$  0.34 \\
\hline

\noalign{\smallskip}
\multirow{3}{*}{Cars}   
&   0.1 & \textbf{22.62} $\pm$  0.14 & \textbf{52.44} $\pm$  0.49 & \textbf{72.99} $\pm$  0.11 & \textbf{87.10} $\pm$  0.20 \\
&  0.01 & 12.10 $\pm$  0.19 & 38.02 $\pm$  0.72 & 63.60 $\pm$  0.10 & 83.20 $\pm$  0.45 \\
& 0.001 &  1.99 $\pm$  0.22 &  3.28 $\pm$  0.12 &  5.16 $\pm$  0.19 & 11.79 $\pm$  0.18 \\
\hline

\noalign{\smallskip}
\multirow{3}{*}{Aircraft}   
&   0.1 & \textbf{24.21} $\pm$  0.60 & \textbf{46.34} $\pm$  0.60 & \textbf{61.70} $\pm$  0.51 & \textbf{78.98} $\pm$  0.28 \\
&  0.01 & 20.02 $\pm$  0.30 & 43.77 $\pm$  0.57 & 61.26 $\pm$  0.27 & 76.53 $\pm$  0.18 \\
& 0.001 &  4.49 $\pm$  0.62 &  7.84 $\pm$  0.62 & 11.94 $\pm$  0.59 & 23.31 $\pm$  0.46 \\
\hline

\noalign{\smallskip}
\multirow{3}{*}{Flowers}   
&   0.1 & \textbf{59.27} $\pm$  0.52 & \textbf{74.86} $\pm$  0.35 & \textbf{82.96} $\pm$  0.35 & \textbf{92.14} $\pm$  0.21 \\
&  0.01 & 40.28 $\pm$  0.81 & 64.22 $\pm$  0.82 & 76.84 $\pm$  0.13 & 87.72 $\pm$  0.05 \\
& 0.001 & 4.23 $\pm$  0.60 & 15.76 $\pm$   1.29 & 30.44 $\pm$  1.98 & 45.72 $\pm$  0.59 \\
\hline

\end{tabular}
\end{center}
\end{table}

\section{Performance according the choice of $\beta$ in SBR}
The appropriate scale of $\beta$ can vary depending on the source and the target task. Table \ref{table:beta_exp_imagenet} and \ref{table:beta_exp_places365} show the variance of the performance according to the choice of $\beta$. Generally, larger $\beta$ works better when using Places365 as a source than ImageNet.

\begin{table}[!h]
\begin{center}
\caption{Test accuracy (\%) of our method with using ImageNet as a source according to $\beta$}
\label{table:beta_exp_imagenet}
\begin{tabular}{C{1.5cm}C{1.5cm}R{2cm}R{2cm}R{2cm}R{2cm}}
\hline\noalign{\smallskip}
\multirow{2}{*}{Dataset} & \multirow{2}{*}{$\beta$} & \multicolumn{4}{c}{Sampling Rate}\\
            \cmidrule(l){3-6}    &  & \multicolumn{1}{c}{15\%} & \multicolumn{1}{c}{30\%} & \multicolumn{1}{c}{50\%} & \multicolumn{1}{c}{100\%} \\
\noalign{\smallskip}
\hline
\hline
\noalign{\smallskip}
\multirow{3}{*}{CUB-200}   
& 1.00E-04 & \textbf{60.11} $\pm$ 0.67 & \textbf{71.50} $\pm$ 0.24 & \textbf{76.91} $\pm$ 0.19 & 82.92 $\pm$ 0.15 \\
& 3.16E-05 & 58.73 $\pm$ 0.13 & 70.86 $\pm$ 0.43 & 76.91 $\pm$ 0.28 & \textbf{83.13} $\pm$ 0.04 \\
& 1.00E-05 & 53.60 $\pm$ 0.29 & 68.11 $\pm$ 0.13 & 75.24 $\pm$ 0.27 & 82.08 $\pm$ 0.21 \\
\hline

\noalign{\smallskip}
\multirow{3}{*}{Dogs}   
& 1.00E-04 & 83.80 $\pm$ 0.12 & 85.36 $\pm$ 0.18 & 86.27 $\pm$ 0.22 & 87.44 $\pm$ 0.15 \\
& 3.16E-05 & \textbf{85.56} $\pm$ 0.14 & 86.90 $\pm$ 0.06 & 87.27 $\pm$ 0.06 & 88.22 $\pm$ 0.03 \\
& 1.00E-05 & 85.50 $\pm$ 0.02 & \textbf{87.44} $\pm$ 0.12 & \textbf{88.03} $\pm$ 0.04 & \textbf{88.91} $\pm$ 0.10 \\
\hline

\noalign{\smallskip}
\multirow{3}{*}{Cars}   
& 1.00E-04 & \textbf{49.12} $\pm$ 1.11 & \textbf{75.24} $\pm$ 0.66 & \textbf{85.61} $\pm$ 0.26 & \textbf{91.73} $\pm$ 0.10 \\
& 3.16E-05 & 47.05 $\pm$ 3.58 & 74.73 $\pm$ 0.99 & 84.85 $\pm$ 0.17 & 91.42 $\pm$ 0.08 \\
& 1.00E-05 & 46.82 $\pm$ 0.91 & 71.60 $\pm$ 0.30 & 83.29 $\pm$ 0.06 & 91.12 $\pm$ 0.19 \\
\hline

\noalign{\smallskip}
\multirow{3}{*}{Aircraft}   
& 1.00E-04 & \textbf{45.26} $\pm$ 0.44 & \textbf{62.91} $\pm$ 0.28 & 73.16 $\pm$ 0.97 & 83.92 $\pm$ 0.21 \\
& 3.16E-05 & 44.41 $\pm$ 0.53 & 62.50 $\pm$ 0.44 & \textbf{73.53} $\pm$ 0.51 & \textbf{84.19} $\pm$ 0.34 \\
& 1.00E-05 & 41.17 $\pm$ 0.92 & 60.23 $\pm$ 0.35 & 71.69 $\pm$ 0.56 & 82.73 $\pm$ 0.15 \\
\hline

\noalign{\smallskip}
\multirow{3}{*}{Flowers}   
& 1.00E-04 & 74.16 $\pm$ 3.23 & 86.82 $\pm$ 0.76 & 91.41 $\pm$ 0.19 & 96.42 $\pm$ 0.11 \\
& 3.16E-05 & \textbf{78.48} $\pm$ 0.59 & \textbf{88.74} $\pm$ 0.38 & 92.15 $\pm$ 0.54 & 96.89 $\pm$ 0.16 \\
& 1.00E-05 & 78.17 $\pm$ 1.08 & 88.63 $\pm$ 0.66 & \textbf{92.47} $\pm$ 0.18 & \textbf{96.98} $\pm$ 0.18 \\
\hline

\end{tabular}
\end{center}
\end{table}

\begin{table}[!h]
\begin{center}
\caption{Test accuracy (\%) of our method with using Places365 as a source according to $\beta$}
\label{table:beta_exp_places365}
\begin{tabular}{C{1.5cm}C{1.5cm}R{2cm}R{2cm}R{2cm}R{2cm}}
\hline\noalign{\smallskip}
\multirow{2}{*}{Dataset} & \multirow{2}{*}{$\beta$} & \multicolumn{4}{c}{Sampling Rate}\\
            \cmidrule(l){3-6}    &  & \multicolumn{1}{c}{15\%} & \multicolumn{1}{c}{30\%} & \multicolumn{1}{c}{50\%} & \multicolumn{1}{c}{100\%} \\
\noalign{\smallskip}
\hline
\hline
\noalign{\smallskip}
\multirow{3}{*}{CUB-200}   
& 1.00E-03 & 20.10 $\pm$ 2.14 & 44.79 $\pm$ 1.08 & 61.11 $\pm$ 1.11 & \textbf{75.06} $\pm$ 0.69 \\
& 3.16E-04 & \textbf{30.46} $\pm$ 0.22 & \textbf{49.90} $\pm$ 0.40 & \textbf{63.34} $\pm$ 0.15 & 74.91 $\pm$ 0.10 \\
& 1.00E-04 & 30.26 $\pm$ 0.37 & 48.75 $\pm$ 0.32 & 62.14 $\pm$ 0.22 & 74.69 $\pm$ 0.09 \\
\hline

\noalign{\smallskip}
\multirow{3}{*}{Dogs}   
& 1.00E-03 & 29.61 $\pm$ 1.64 & 49.73 $\pm$ 1.46 & 61.60 $\pm$ 0.33 & 71.62 $\pm$ 0.81 \\
& 3.16E-04 & \textbf{39.46} $\pm$ 0.43 & \textbf{53.13} $\pm$ 0.18 & \textbf{62.27} $\pm$ 0.27 & \textbf{71.68} $\pm$ 0.26 \\
& 1.00E-04 & 39.18 $\pm$ 0.45 & 51.79 $\pm$ 0.30 & 60.65 $\pm$ 0.12 & 70.62 $\pm$ 0.62 \\
\hline

\noalign{\smallskip}
\multirow{3}{*}{Cars}   
& 1.00E-03 & 27.67 $\pm$ 2.26 & 65.54 $\pm$ 0.95 & 81.20 $\pm$ 0.40 & 90.40 $\pm$ 0.20 \\
& 3.16E-04 & \textbf{42.76} $\pm$ 0.24 & \textbf{68.76} $\pm$ 0.30 & \textbf{81.98} $\pm$ 0.06 & \textbf{90.59} $\pm$ 0.15 \\
& 1.00E-04 & 40.45 $\pm$ 0.25 & 68.21 $\pm$ 0.42 & 81.54 $\pm$ 0.06 & 90.47 $\pm$ 0.10 \\
\hline

\noalign{\smallskip}
\multirow{3}{*}{Aircraft}   
& 1.00E-03 & 29.81 $\pm$ 2.51 & 58.56 $\pm$ 0.57 & \textbf{72.80} $\pm$ 0.57 & \textbf{86.03} $\pm$ 0.33 \\
& 3.16E-04 & \textbf{37.68} $\pm$ 1.32 & \textbf{59.19} $\pm$ 0.89 & 71.77 $\pm$ 0.68 & 85.04 $\pm$ 0.20 \\
& 1.00E-04 & 36.44 $\pm$ 0.06 & 57.63 $\pm$ 0.26 & 70.78 $\pm$ 0.03 & 83.91 $\pm$ 0.66 \\
\hline

\noalign{\smallskip}
\multirow{3}{*}{Flowers}   
& 3.16E-04 & 64.96 $\pm$ 0.49 & 79.78 $\pm$ 0.64 & 86.91 $\pm$ 0.39 & 94.18 $\pm$ 0.12 \\
& 1.00E-04 & \textbf{67.47} $\pm$ 0.61 & \textbf{82.26} $\pm$ 0.69 & 88.19 $\pm$ 0.26 & 94.73 $\pm$ 0.15 \\
& 3.16E-05 & 64.79 $\pm$ 0.51 & 81.51 $\pm$ 0.67 & \textbf{88.41} $\pm$ 0.07 & \textbf{94.94} $\pm$ 0.10 \\
\hline

\end{tabular}
\end{center}
\end{table}

\section{Comparison results when using Places365 as a source}
To validate the generality of our method regardless of the choice of the source model, we examined additional experiments using Places365 as a source instead of using ImageNet. Table \ref{table:main_exp_Places365} shows the results of various configurations; this is the raw result of Fig. 2 in the main paper. Our method consistently outperforms other methods with a healthy margin.

\begin{table}[h!]
\begin{center}
\caption{Test accuracy (\%) of various methods when using Places365 as a source}
\label{table:main_exp_Places365}
\begin{tabular}{C{1.5cm}C{2cm}R{2cm}R{2cm}R{2cm}R{2cm}}
\hline\noalign{\smallskip}
\multirow{2}{*}{Dataset} & \multirow{2}{*}{Method} & \multicolumn{4}{c}{Sampling Rate}\\
            \cmidrule(l){3-6}    &  & \multicolumn{1}{c}{15\%} & \multicolumn{1}{c}{30\%} & \multicolumn{1}{c}{50\%} & \multicolumn{1}{c}{100\%} \\
\noalign{\smallskip}
\hline
\noalign{\smallskip}
\multirow{7}{*}{CUB}   
&  Baseline  &15.08$\pm$0.16 & 31.65$\pm$0.76 & 48.56$\pm$0.59 & 66.43$\pm$0.35 \\
&  L2-SP     &15.09$\pm$0.34 &31.89$\pm$0.58 &48.45$\pm$0.49 &67.39$\pm$0.43 \\
&  DELTA     &14.63$\pm$0.23 &31.75$\pm$0.50 &48.31$\pm$0.94 &66.97$\pm$0.10 \\
&  BSS+L2     &16.65$\pm$0.19 &35.41$\pm$0.30 &52.36$\pm$0.61 &68.57$\pm$0.37 \\
&  BSS+L2-SP  &16.95$\pm$0.12 &35.07$\pm$0.59 &52.16$\pm$0.47 &69.44$\pm$0.07 \\
&  BSS+DELTA  &16.76$\pm$0.31 &35.48$\pm$0.42 &52.64$\pm$0.45 &68.84$\pm$0.58 \\
&  SBR(ours) &\textbf{30.46}$\pm$0.22 & \textbf{49.90}$\pm$0.40 & \textbf{63.34}$\pm$0.15 & \textbf{74.91}$\pm$0.10 \\
\hline

\noalign{\smallskip}
\multirow{7}{*}{Dogs}
&  Baseline  &32.13$\pm$0.27 & 44.92$\pm$0.07 & 55.14$\pm$0.38 & 66.34$\pm$0.15 \\
&  L2-SP     &32.40$\pm$0.22 &45.75$\pm$0.30 &55.63$\pm$0.16 &66.70$\pm$0.32 \\
&  DELTA     &31.97$\pm$0.21 &45.79$\pm$0.32 &55.34$\pm$0.15 &66.23$\pm$0.06 \\
&  BSS+L2    &33.54$\pm$0.09 &47.17$\pm$0.22 &56.94$\pm$0.39 &67.67$\pm$0.37 \\
&  BSS+L2-SP &33.50$\pm$0.22 &47.45$\pm$0.41 &57.34$\pm$0.20 &67.82$\pm$0.11 \\
&  BSS+DELTA &33.60$\pm$0.29 &46.97$\pm$0.14 &56.27$\pm$0.18 &66.76$\pm$0.19 \\
&  SBR(ours) &\textbf{39.46}$\pm$0.43 & \textbf{53.13}$\pm$0.18 & \textbf{62.27}$\pm$0.27 & \textbf{71.68}$\pm$0.26 \\
\hline

\noalign{\smallskip}
\multirow{7}{*}{Cars}
&  Baseline  &22.62$\pm$0.14 & 52.44$\pm$0.49 & 72.99$\pm$0.11 & 87.10$\pm$0.20 \\
&  L2-SP     &22.75$\pm$0.38 &52.32$\pm$0.67 &73.05$\pm$0.24 &87.58$\pm$0.08 \\
&  DELTA     &22.66$\pm$0.45 &52.27$\pm$0.27 &72.95$\pm$0.70 &87.17$\pm$0.24 \\
&  BSS+L2     &22.92$\pm$0.39 &55.71$\pm$0.61 &75.99$\pm$0.32 &88.61$\pm$0.09 \\
&  BSS+L2-SP  &23.56$\pm$1.07 &55.08$\pm$3.00 &76.55$\pm$1.56 &88.58$\pm$0.22 \\
&  BSS+DELTA  &21.48$\pm$2.64 &56.71$\pm$0.62 &75.94$\pm$1.38 &88.11$\pm$0.76 \\
&  SBR(ours) &\textbf{42.76}$\pm$0.24 & \textbf{68.76}$\pm$0.30 & \textbf{81.98}$\pm$0.06 & \textbf{90.59}$\pm$0.15 \\
\hline

\noalign{\smallskip}
\multirow{7}{*}{Aircraft}
&  Baseline  &24.21$\pm$0.60 & 46.34$\pm$0.60 & 61.70$\pm$0.51 & 78.98$\pm$0.28 \\
&  L2-SP     &24.55$\pm$0.18 &46.65$\pm$0.52 &62.38$\pm$0.21 &79.49$\pm$0.36 \\
&  DELTA     &24.52$\pm$0.81 &46.16$\pm$0.55 &61.86$\pm$0.82 &78.79$\pm$0.34 \\
&  BSS+L2     &26.35$\pm$0.83 &48.27$\pm$0.82 &63.42$\pm$0.24 &79.92$\pm$0.48 \\
&  BSS+L2-SP  &25.67$\pm$0.29 &48.89$\pm$0.33 &65.07$\pm$0.36 &81.16$\pm$0.36 \\
&  BSS+DELTA  &25.96$\pm$0.64 &49.06$\pm$0.80 &65.22$\pm$0.44 &81.05$\pm$0.39 \\
&  SBR(ours) &\textbf{37.68}$\pm$1.32 & \textbf{59.19}$\pm$0.89 & \textbf{71.77}$\pm$0.68 & \textbf{85.04}$\pm$0.20 \\
\hline

\noalign{\smallskip}
\multirow{7}{*}{Flowers}
&  Baseline  &59.27$\pm$0.52 & 74.86$\pm$0.35 & 82.96$\pm$0.35 & 92.14$\pm$0.21 \\
&  L2-SP     &60.34$\pm$0.29 &75.33$\pm$0.65 &82.90$\pm$0.30 &92.44$\pm$0.27 \\
&  DELTA     &59.83$\pm$0.60 &75.76$\pm$0.45 &83.05$\pm$0.72 &92.14$\pm$0.34 \\
&  BSS+L2     &59.63$\pm$0.26 &75.31$\pm$0.29 &83.49$\pm$0.21 &92.37$\pm$0.36 \\
&  BSS+L2-SP  &59.15$\pm$1.15 &75.23$\pm$0.26 &83.16$\pm$0.50 &92.54$\pm$0.20 \\
&  BSS+DELTA  &59.22$\pm$0.05 &74.97$\pm$0.39 &83.46$\pm$0.20 &92.75$\pm$0.32 \\
&  SBR(ours) &\textbf{67.47}$\pm$0.61 & \textbf{82.26}$\pm$0.69 & \textbf{88.19}$\pm$0.26 & \textbf{94.73}$\pm$0.15 \\
\hline

\end{tabular}
\end{center}
\end{table}
